\documentclass[final]{cvpr}

\usepackage{times}
\usepackage{epsfig}
\usepackage{graphicx}
\usepackage{amsmath}
\usepackage{amssymb}
\usepackage{caption}
\usepackage[ruled,vlined]{algorithm2e}

\usepackage{subfigure}
\usepackage{subfloat}
\usepackage{multirow}

\usepackage{anyfontsize}
\usepackage{makecell}
\usepackage{commath}
\usepackage{textcomp}

\newcommand{\OM}{VRCNet}

\usepackage[symbol]{footmisc}

\usepackage[pagebackref=true,breaklinks=true,colorlinks,bookmarks=false]{hyperref}
\usepackage{marvosym}

\def\cvprPaperID{2755} 
\def\confYear{CVPR 2021}
\begin{document}

\title{Variational Relational Point Completion Network}

\author{Liang Pan$^{1\star}$ \quad
Xinyi Chen$^{1,2}$ \quad
Zhongang Cai$^{2,3}$ \quad
Junzhe Zhang$^{1,2}$ \quad
\\Haiyu Zhao$^{2,3}$ \quad
Shuai Yi$^{2,3}$ \quad
Ziwei Liu$^{1\href{mailto:ziwei.liu@ntu.edu.sg}{\textrm{\Letter}}}$ \\
$^{1}$S-Lab, Nanyang Technological University \hspace{6pt} $^{2}$SenseTime Research \hspace{6pt} $^{3}$Shanghai AI Laboratory \\
{\tt\small {\{liang.pan,ziwei.liu\}@ntu.edu.sg}, \{xchen032,junzhe001\}@e.ntu.edu.sg} \\
{\tt\small {\{caizhongang,zhaohaiyu,yishuai\}@sensetime.com}} \\
}


\twocolumn[{
    \renewcommand\twocolumn[1][]{#1}%
    \maketitle
    \vspace{-38pt}
    \begin{center}
        \centering
        \includegraphics[width=1\textwidth]{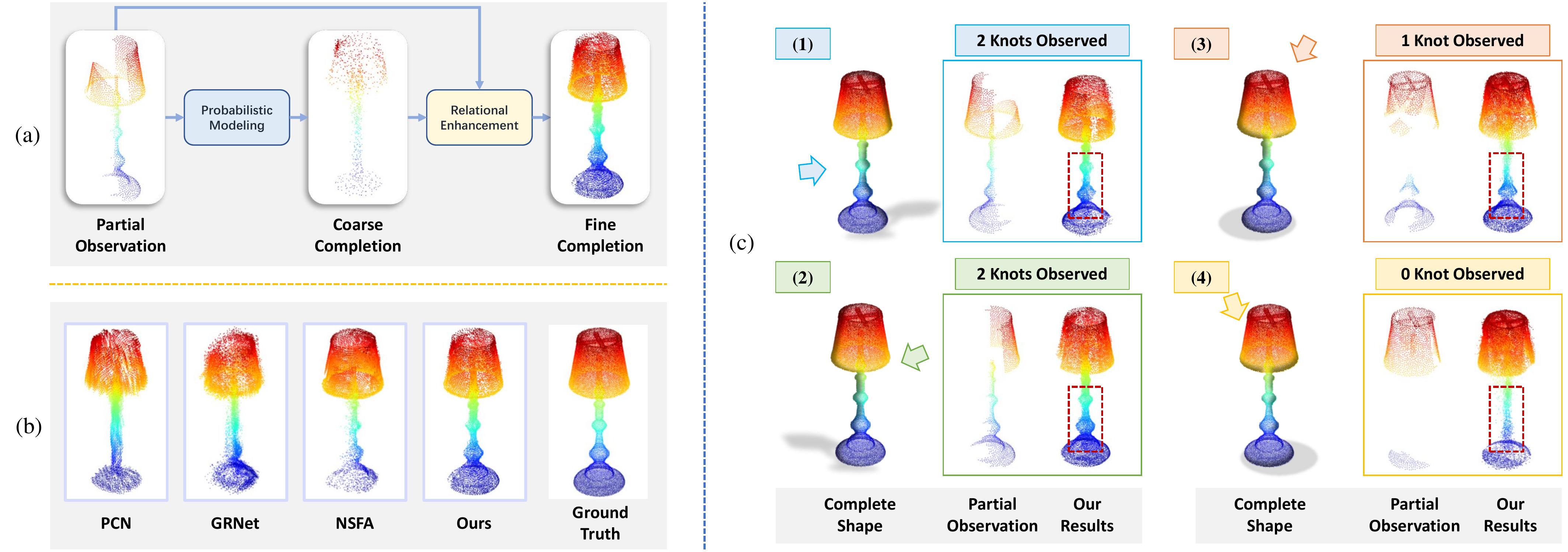}
        \vspace{-8mm}
        \captionof{figure}{(a) 
        \textbf{ 
        \OM{} performs shape completion
        with two consecutive stages}: probabilistic modeling and relational enhancement.
        (b) 
        \textbf{Qualitative Results} 
        show that 
        \OM{}
        generates better 
        shape
        details than 
        the other works~\cite{yuan2018pcn,xie2020grnet,zhang2020detail}.
        (c) \textbf{Our completion results conditioned on partial observations.} 
        The arrows indicate the viewing angles. 
        In (1) and (2), 
        2 knots are partially observed for the pole of the lamp, and hence we generate 2 complete knots.
        In (3), 
        only 1 knot is observed, and then we reconstruct 1 complete knot. 
        If no knots are observed (see (4)), \OM{} generates a smooth pole without knots.
        }
        \label{fig:first_glance}
    \end{center}
}]

\begin{abstract}
   Real-scanned point clouds are often incomplete due to viewpoint, occlusion, and noise.
   Existing point cloud completion methods tend to generate 
   global shape skeletons
   and hence lack fine local details.
   Furthermore, they
   mostly learn a deterministic partial-to-complete mapping, but 
   overlook structural relations
   in man-made objects.
   To tackle these challenges, this paper proposes a variational framework, \textbf{V}ariational \textbf{R}elational point \textbf{C}ompletion network (\OM{}) with two appealing properties:
   \textbf{1) Probabilistic Modeling.} In particular, we propose a dual-path architecture to enable principled probabilistic modeling across partial and complete clouds.
   One path consumes complete point clouds for reconstruction by learning a point VAE.
   The other path generates complete shapes for partial point clouds, whose embedded distribution is guided by distribution obtained 
    from
   the reconstruction path during training.
   \textbf{2) Relational Enhancement.} Specifically, we carefully design point self-attention kernel and point selective kernel module to exploit relational point features, which refines local shape details conditioned on the coarse completion.
   In addition, we contribute \textbf{a multi-view partial point cloud dataset (MVP dataset)} containing over 100,000 high-quality scans, which renders partial 3D shapes from 26 uniformly distributed camera poses 
   for each 3D CAD model.
   Extensive experiments demonstrate that \OM{} outperforms state-of-the-art methods on all standard point cloud completion benchmarks. 
   Notably, \OM{} shows great generalizability and robustness on real-world point cloud scans.
  \footnote[0]{
$^\star$ Work partially done while working at NUS.
  }
  \footnote[0]{
  Our project website: \href{https://paul007pl.github.io/projects/VRCNet}{https://paul007pl.github.io/projects/VRCNet}
  }
\end{abstract}

\section{Introduction}

3D point cloud is an intuitive representation of 3D scenes and objects, which has extensive applications in various vision and robotics tasks.
Unfortunately,
scanned 3D point clouds are usually incomplete owing to occlusions and 
missing
measurements, 
hampering practical usages.
Therefore,
it is desirable and important to predict the complete 3D shape from a partially observed point cloud.


The pioneering work PCN~\cite{yuan2018pcn} 
uses PointNet-based encoder to generate global features for shape completion,
which cannot recover fine geometric details.
The follow-up works~\cite{liu2020morphing,wang2020cascaded,pan2020ecg,xie2020grnet} provide better completion results by preserving observed geometric details from the incomplete point shape using local features.
However, 
they~\cite{yuan2018pcn,liu2020morphing,wang2020cascaded,pan2020ecg,xie2020grnet} mostly generate complete shapes by learning a deterministic partial-to-complete mapping, lacking the conditional generative capability 
based on the partial observation.
Furthermore, 
3D shape completion is expected to recover plausible yet fine-grained complete shapes by learning relational structure properties, such as
geometrical symmetries, regular arrangements and surface smoothness,
which existing methods fail to capture.


To this end,
we propose Variational Relational Point Completion network (entitled as \OM), 
which consists of two consecutive encoder-decoder sub-networks that serve as 
``probabilistic modeling'' (PMNet)
and 
``relational enhancement'' (RENet), respectively
(shown in Fig.~\ref{fig:first_glance}~(a)).
The first sub-network, PMNet, 
embeds global features and latent distributions from incomplete point clouds, and predicts 
the
overall skeletons (\ie coarse completions, see Fig.~\ref{fig:first_glance} (a))
that are used as 3D adaptive anchor points for exploiting multi-scale point relations in RENet.
Inspired by~\cite{zheng2019pluralistic},
PMNet 
uses smooth complete shape priors
to improve the generated 
coarse completions
using a dual-path 
architecture
consisting
of two parallel paths: 1) a reconstruction path for complete point clouds, and 2) a completion path for incomplete point clouds.
During training, we 
regularize the consistency
between the encoded posterior distributions from partial point clouds and the prior distributions from complete point clouds.
With the help of the generated 
coarse completions,
the second sub-network 
RENet strives to 
enhance structural relations
by learning multi-scale local point features.
Motivated by the success of local 
relation operations
in image recognition~\cite{zhao2020exploring,hu2019local}, 
we propose the Point Self-Attention Kernel (PSA) as a basic building block for 
RENet.
Instead of 
using fixed 
weights,
PSA
interleaves local point features by adaptively predicting weights 
based on the learned relations
among neighboring points.
Inspired by the Selective Kernel (SK) unit~\cite{li2019selective}, we propose the Point Selective Kernel Module (PSK) that utilizes multiple branches with different kernel sizes to exploit and fuse multi-scale point features, which further improves the performance.

Moreover,
we create a large-scale Multi-View Partial point cloud (MVP) dataset with over 100,000 high-quality scanned partial and complete point clouds.
For each complete 3D CAD model selected from ShapeNet~\cite{wu20153d}, we randomly render 26 partial point clouds from uniformly distributed camera views on a unit sphere, which improves the data diversity.
Experimental results 
on our MVP and Completion3D benchmark~\cite{tchapmi2019topnet}
show that 
\OM{} 
outperforms 
SOTA methods.
In Fig.~\ref{fig:first_glance} (b), \OM{} 
reconstructs richer 
details than the other methods 
by implicitly learning the shape symmetry 
from this incomplete lamp.
Given different partial observations, \OM{} can predict different plausible complete shapes 
(Fig.~\ref{fig:first_glance} (c)).
Furthermore, \OM{}
can generate impressive complete shapes for 
incomplete real-world scans from KITTI~\cite{Geiger2012CVPR} and ScanNet~\cite{dai2017scannet}, which reveals its remarkable robustness and generalizability.

The key contributions can be summarized as:
\textbf{1)} We propose a novel \textbf{V}ariational \textbf{R}elational point \textbf{C}ompletion \textbf{Net}work (\OM), and it first performs probabilistic modeling using a novel dual-path network followed by a relational enhancement network.
\textbf{2)} We design multiple relational modules that can effectively exploit and fuse multi-scale point features for point cloud analysis, such as the Point Self-Attention Kernel and the Point Selective Kernel Module.
\textbf{3)} Furthermore, we contribute a large-scale multi-view partial point cloud (MVP) dataset with over 100,000 high-quality 3D point shapes. Extensive experiments show that \OM{} outperforms previous SOTA methods on all evaluated benchmark datasets.

\begin{figure*}
    \centering
    \includegraphics[width=.95\linewidth]{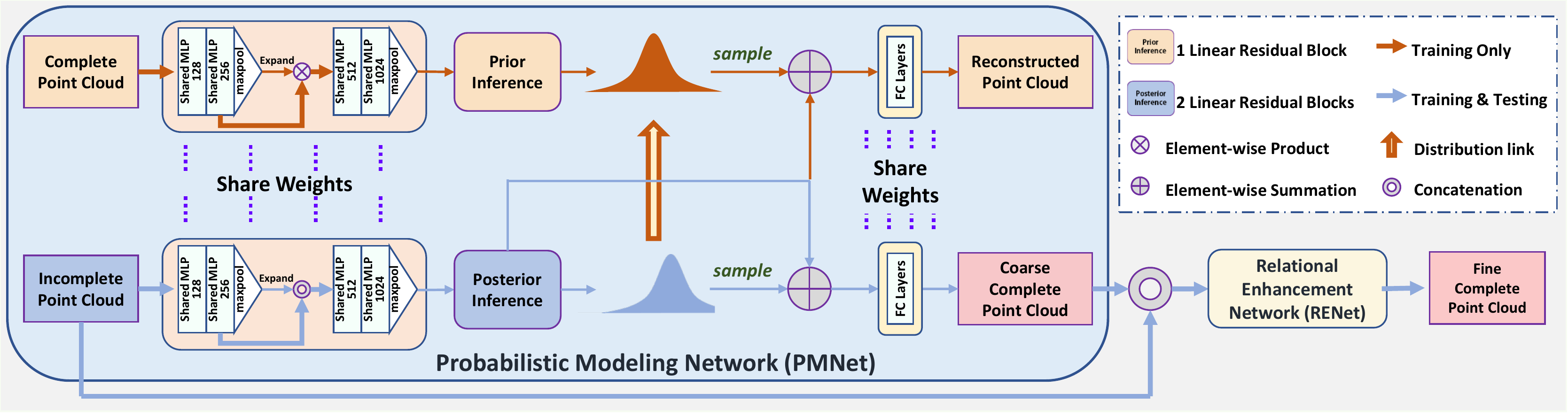}
    \vspace{-3.5mm}
    \caption{
    \textbf{
    Our PMNet (light blue block) consists of two parallel paths}, the upper construction path (orange line) and the lower completion path (blue line). The reconstruction path is only used in training, and the completion path generates a coarse complete point cloud 
    based on the inferred distribution and global features. Subsequently, our RENet (Fig \ref{fig:spsakn}) adaptively exploits relational structure properties to predict the fine complete point cloud.}
    \label{fig:overview}
\end{figure*}
\section{Related Works}

\noindent\textbf{Multi-scale Features Exploitation.}
Convolutional operations have yielded impressive results for image applications 
~\cite{krizhevsky2017imagenet,he2016deep,simonyan2014very}.
However, 
conventional convolutions cannot be directly applied to point clouds due to the absence of regular grids.
Previous 
networks mostly exploit
local point features 
by two operations:
local pooling~\cite{wang2019dynamic,qi2017pointnet++,pan2019pointatrousgraph}
and
flexible convolution~\cite{groh2018flex,thomas2019kpconv,li2018pointcnn,wu2019pointconv}.
Self-attention often uses linear layers, such as fully-connected (FC) layers and shared multilayer perceptron (shared MLP) layers, which are appropriate for point clouds.
In particular, recent works~\cite{zhao2020exploring,hu2019local,parmar2019stand} have shown that local self-attention (\ie relation operations) can outperform their convolutional counterparts, which holds the exciting prospect of designing networks for point clouds.


\noindent\textbf{Point Cloud Completion.}
The target of point cloud completion is to recover a complete 3D shape based on a partial point cloud observation.
PCN~\cite{yuan2018pcn} 
first generates a coarse completion based on learned global features from the partial input point cloud, which is upsampled using folding operations~\cite{yang2018foldingnet}.
Following PCN, TopNet~\cite{tchapmi2019topnet} proposes a tree-structured decoder to predict complete shapes.
To preserve and recover local details, previous approaches~\cite{wang2020cascaded,pan2020ecg,xie2020grnet} exploit local features to refine their 3D completion results.
Recently, NSFA~\cite{zhang2020detail} recovers complete 3D shapes by combining known features and missing features.
However, NSFA assumes that the ratio of the known part and the missing part is around $1:1$ (\ie, the visible part should be roughly a half of the whole object), which does not hold for point clouds completion in most cases.

\section{Our Approach}
We define 
the incomplete point cloud $\mathbf{X}$ 
as
a partial observation for a 3D object,
and a complete point cloud $\mathbf{Y}$ is sampled from the surfaces of the object. 
Note that $\mathbf{X}$ need not to be a subset of $\mathbf{Y}$, since $\mathbf{X}$ and $\mathbf{Y}$ are generated by two separate processes.
The point cloud completion task aims to predict a complete shape $\mathbf{Y}^{\prime}$ conditioned on 
$\mathbf{X}$.
\OM{} generate high-quality complete point clouds in a coarse-to-fine fashion.
Firstly, we predict a coarse completion $\mathbf{Y}^{\prime}_{c}$ based on embedded global features and an estimated parametric distribution.
Subsequently, we 
recover relational geometries
for the fine completion $\mathbf{Y}^{\prime}_{f}$ by 
exploiting multi-scale point features
with novel self-attention modules.

\begin{figure*}
    \centering
    \includegraphics[width=0.95\linewidth]{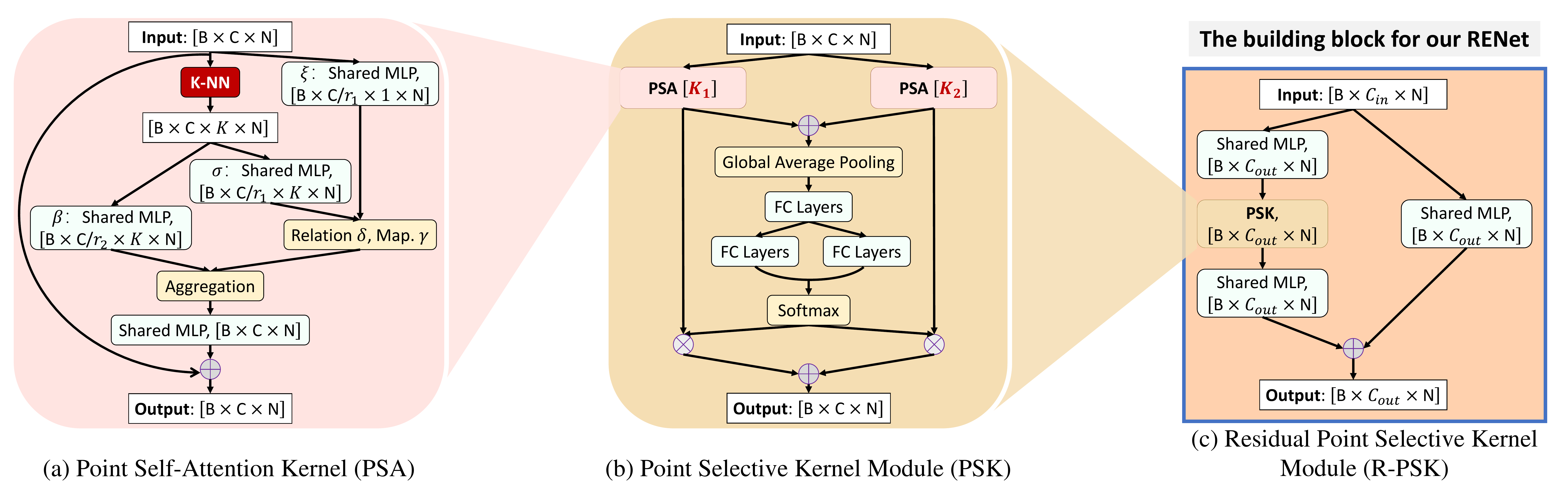}
    \vspace{-5mm}
    \caption{\textbf{Our proposed point kernels.} (a) Our PSA adaptively aggregate neighboring point features. (b) Using selective kernel unit, our PSK can adaptively adjust receptive fields to exploit and fuse multi-scale point features. (c) By adding a residual connection, we construct our RPSK that is an important building block for our RENet.}
    \vspace{-2mm}
    \label{fig:point_modules}
\end{figure*}

\subsection{Probabilistic Modeling}
Previous networks~\cite{yuan2018pcn,tchapmi2019topnet}
tend to 
decode learned global features
to
predict 
overall shape skeletons 
as their completion results,
which
cannot recover fine-grained geometric details.
However,
it is still beneficial to first predict 
the shape skeletons
before refining local details for the following reasons:
1) 
shape skeletons
describe the coarse complete structures,
especially for 
those areas that are entirely missing in the 
partial observations;
2) 
shape skeletons
can be regarded as adaptive 3D anchor points for 
exploiting local point features in incomplete point clouds
~\cite{pan2020ecg}.
With these benefits, 
we propose the
Probabilistic Modeling network (PMNet) to generate 
the overall skeletons (\ie coarse completions)
for incomplete point clouds.

In contrast to previous methods,
PMNet employs probabilistic modeling to predict 
the coarse completions
based on both embedded global features and learned latent distributions.
Moreover, we employ a dual-path architecture (shown in Fig.~\ref{fig:overview}) that contains two parallel pipelines: the upper reconstruction path for complete point clouds $\mathbf{Y}$ and the lower completion path for partial point clouds $\mathbf{X}$.
The reconstruction path follows a variational auto-encoder (VAE) scheme. 
It first encodes global features $\mathbf{z_g}$ and latent distributions $\,q_{\phi}(\mathbf{z_g}|\mathbf{Y})$ for the complete shape $\mathbf{Y}$, and then it uses a decoding distribution $p_{\theta}^r(\mathbf{Y}|\mathbf{z_g})$ to recover a complete shape $\mathbf{Y}_r^{\prime}$.
The objective function for the reconstruction path can be formulated as:
\begin{equation}
    \begin{aligned}
        \mathcal{L}_{rec} = &-\lambda \, \mathbf{KL}\big[q_{\phi}(\mathbf{z_g}|\mathbf{Y}) \, \big\| \,p(\mathbf{z_g})\big] \\
        &+ \mathbb{E}_{p_{data}(\mathbf{Y})}\mathbb{E}_{q_{\phi}(\mathbf{z_g}|\mathbf{Y})}\big[\log {p_{\theta}^r}(\mathbf{Y}|\mathbf{z_g})\big],
    \end{aligned}
\end{equation}
where 
$\mathbf{KL}$
is the KL divergence, 
$\mathbb{E}$ denotes the estimated expectations of certain functions,
$p_{data}(\mathbf{Y})$ 
denotes the true underlying distribution of data,
and $p(\mathbf{z_g})=\mathcal{N}(\mathbf{0}, \mathbf{I})$ is the conditional prior predefined as a Gaussian distribution, and $\lambda$ is a weighting parameter.

The completion path has a similar structure as the constructive path, and both two paths share weights for their encoder and decoder except the distribution inference layers.
Likewise, the completion path aims to reconstruct a complete shape $\mathbf{Y}^{\prime}_c$ based on global features $\mathbf{z_g}$ and latent distributions $p_{\psi}(\mathbf{z_g}|\mathbf{X})$ from an incomplete input $\mathbf{X}$.
To exploit the most salient features from the incomplete point cloud, we use the learned conditional distribution $q_{\phi}(\mathbf{z_g}|\mathbf{Y})$ encoded by its corresponding complete 3D shapes $\mathbf{Y}$ to 
regularize
latent distributions $p_{\psi}(\mathbf{z_g}|\mathbf{X})$ during training (shown as the Distribution Link in Fig.~\ref{fig:overview}, the arrow indicates that we regularize $p_{\psi}(\mathbf{z_g}|\mathbf{X})$ to approach $q_{\phi}(\mathbf{z_g}|\mathbf{Y})$).
Hence, $q_{\phi}(\mathbf{z_g}|\mathbf{Y})$ constitutes the prior distributions, $p_{\psi}(\mathbf{z_g}|\mathbf{X})$ is the posterior importance sampling function, and the objective function for completion path is defined as follows:
\begin{equation}
    \begin{aligned}
        \mathcal{L}_{com} = &-\lambda \, \mathbf{KL}\big[q_{\phi}(\mathbf{z_g}|\mathbf{Y}) \, \big\| \,p_{\psi}(\mathbf{z_g}|\mathbf{X})\big] \\
        &+ 
        \mathbb{E}_{p_{data}(\mathbf{X})}\mathbb{E}_{p_{\psi}(\mathbf{z_g}|\mathbf{X})}\big[\log {p_{\theta}^c}(\mathbf{Y}|\mathbf{z_g})\big],
    \end{aligned}
\end{equation}
where $\phi$, $\psi$ and $\theta$ represent different network weights of their corresponding functions.
Notably,
the reconstruction path is only used in training, and hence the dual-path architecture does not influence our inference efficiency.

\subsection{Relational Enhancement}
After 
obtaining
coarse completions
$\mathbf{Y}^{\prime}_c$, the 
Relational Enhancement network (RENet)
targets at
enhancing 
structural relations 
to recover local shape details.
Although previous methods~\cite{wang2020cascaded,zhang2020detail,pan2020ecg} can preserve observed geometric details by exploiting local point features,
they cannot effectively extract structural relations (\eg geometric symmetries) to recover those missing parts conditioned on the partial observations.
Inspired by the
relation operations for image recognition~\cite{hu2019local,zhao2020exploring},
we
propose the Point Self-Attention kernel (PSA) to adaptively aggregate local neighboring point features with learned 
relations in neighboring points
(Fig.~\ref{fig:point_modules} (a)). 
The operation of 
PSA 
is
formulated as:
\begin{equation}
    \mathbf{y}_i = \underset{j\in \mathcal{N}(i)}{\sum}\alpha (\mathbf{x}_{\mathcal{N}(i)})_j \, \odot \beta(\mathbf{x}_j),
\end{equation}
where $\mathbf{x}_{\mathcal{N}(i)}$ is the group of point feature vectors for the selected K-Nearest Neighboring (K-NN) points $\mathcal{N}(i)$.
$\alpha (\mathbf{x}_{\mathcal{N}(i)})$ is a weighting tensor for all selected feature vectors.
$\beta(\mathbf{x}_j)$ is the transformed features for point $j$, which has the same spatial dimensionality with $\alpha (\mathbf{x}_{\mathcal{N}(i)})_j$.
Afterwards,
we 
obtain the output $\mathbf{y}_i$ using
an element-wise product $\odot$,
which performs
a weighted summation
for all points $j\in \mathcal{N}(i)$.
The weight computation $\alpha (\mathbf{x}_{\mathcal{N}(i)})$ can be decomposed as follows:
\begin{equation}
    \begin{aligned}
        \alpha (\mathbf{x}_{\mathcal{N}(i)}) &= \gamma\big(\delta(\mathbf{x}_{\mathcal{N}(i)})\big), \\
        \delta(\mathbf{x}_{\mathcal{N}(i)}) &= 
        \big[\sigma(\mathbf{x}_i), [\xi(\mathbf{x}_j)]_{\forall j \in \mathcal{N}(i)}\big],
    \end{aligned}
\end{equation}
where $\gamma$, $\sigma$ and $\xi$ are all shared MLP layers (Fig.~\ref{fig:point_modules} (a)), and the relation function $\delta$ 
combines all feature vectors $\mathbf{x}_j \in \mathbf{x}_{\mathcal{N}(i)}$ by using concatenation operations.

\begin{figure}
    \centering
    \includegraphics[width=1\linewidth]{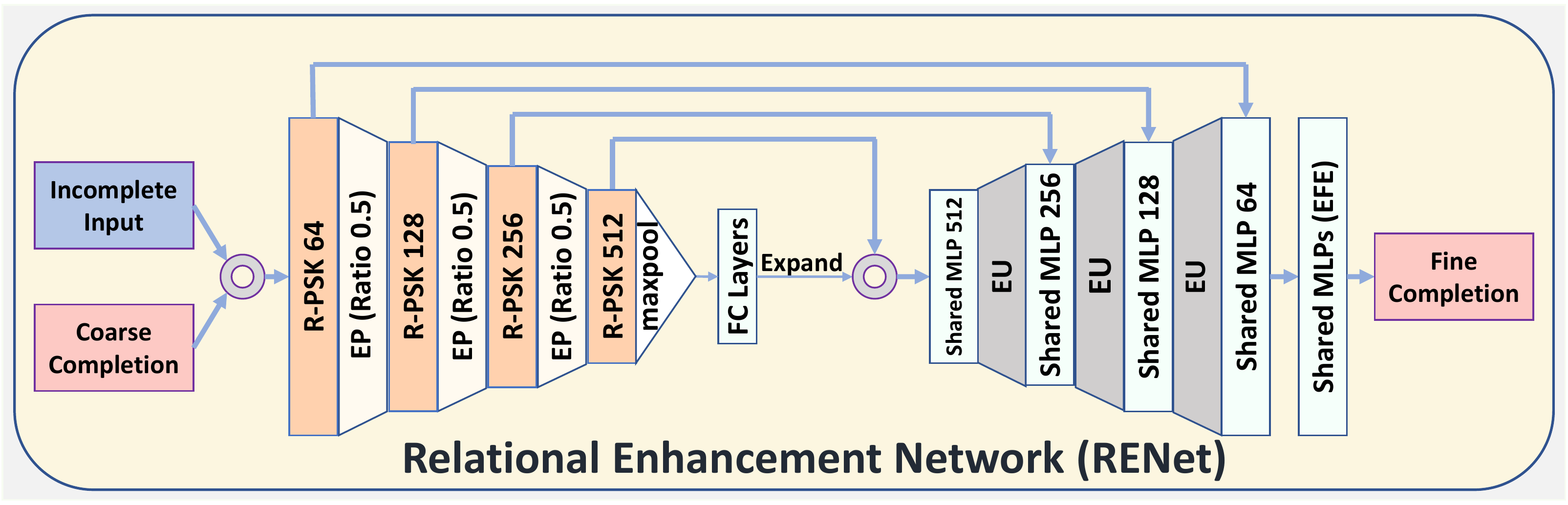}
    \vspace{-8mm}
    \caption{\textbf{Our Relational Enhancement Network (RENet)} uses a hierarchical encoder-decoder architecture, which effectively learns multi-scale structural relations.}
    \label{fig:spsakn}
\end{figure}
\setlength{\tabcolsep}{4.85pt}
\begin{table*}[]
    \caption{\textbf{Comparing MVP with existing datasets.} 
    MVP has many appealing properties, such as
    1) diversity of uniform views; 2) large-scale and high-quality; 3) rich categories. 
    Note that both PCN and C3D only randomly render \textbf{One} incomplete point cloud for each CAD model to construct their testing sets.
    (C3D: Completion3D; Cat.: Categories; Distri.: Distribution; Reso.: Resolution; PC: Point Cloud; FPS: Farthest Point Sampling; PDS: Poisson Disk Sampling. Point cloud resolution is shown as multiples of 2048 points.)}
    \vspace{-7mm}
    \begin{center}
    \small{
        \begin{tabular}{l|c|cc|cc|ccc|cc|cc}
            \Xhline{1pt}
             \multirow{2}{*}{} &
             \multirow{2}{*}{\#Cat.} & 
             \multicolumn{2}{c|}{Training Set} &
             \multicolumn{2}{c}{Testing Set} &
             \multicolumn{3}{|c|}{Virtual Camera} & 
             \multicolumn{2}{c|}{Complete PC} &
             \multicolumn{2}{c}{Incomplete PC} \\
             & & \#CAD & \#Pair & \#CAD & \#Pair & Num. & Distri. & Reso. & 
             Sampling & Reso. & Sampling & Reso. \\
            \hline\hline 
             \small PCN~\cite{yuan2018pcn} & \small 8 & \small 28974 & \small $\sim$200k & \small 1200 & \small 1200 & \small 8 & \small Random & \small {160\texttimes120} & \small Uniform & \small 8\texttimes & \small Random & \small $\sim$3000 \\ 
             \small C3D~\cite{tchapmi2019topnet} & \small 8 & 28974 & \small 28974 & \small 1184 & \small 1184 & \small 1 & \small Random & \small {160\texttimes120} & \small Uniform & \small 1\texttimes & \small Random & \small 1\texttimes \\ 
            \hline
             MVP & \small \textbf{16} & \small 2400 & \small  62400  & \small 1600 & \textbf{41600} & \small \textbf{26} & \small \textbf{Uniform} & \small \textbf{{1600\texttimes1200}} & \small \textbf{PDS} & \small \textbf{1/2/4/8\texttimes} & 
             \small \textbf{FPS} & \small 1\texttimes \\
            \Xhline{1pt}
        \end{tabular}
    }
    \end{center}
    \vspace{-3mm}
    \label{tab:mvp_comp}
\end{table*}

\begin{figure*}
    \centering
    \vspace{-4.5mm}
    \includegraphics[width=1\linewidth]{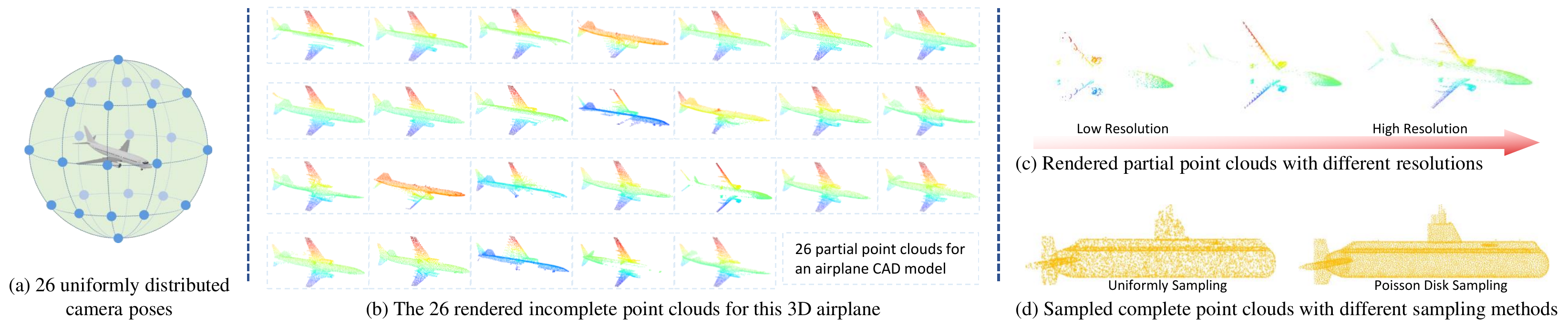}
    \vspace{-8.25mm}
    \caption{Our \textbf{M}ulti-\textbf{V}iew \textbf{P}artial point cloud dataset (\textbf{MVP}). (a) shows an example for our 26 uniformly distributed camera poses on a unit sphere. (b) presents the 26 partial point clouds for the airplane from our uniformly distributed virtual cameras. (c) compares the rendered incomplete point clouds with different camera resolutions. (d) shows that Poisson disk sampling generates complete point clouds with a higher quality than uniform sampling.}
    \vspace{-2mm}
    \label{fig:mv_all}
\end{figure*}

Observing that different relational structures can have 
different scales,
we
enable the neurons to adaptively adjust their receptive field sizes
by leveraging
the selective kernel unit~\cite{li2019selective}. 
Hence, we 
construct the Point Selective Kernel module (PSK), which adaptively fuses learned 
structural relations from different scales.
In Fig.~\ref{fig:point_modules} (b), we show a two-branch case, which has two PSA kernels with different kernel (\ie K-NN) sizes.
The operations of the PSK are formulated as:
\begin{equation}
\hspace{-2mm}
\left\{
    \begin{aligned}
        & \mathbf{V}_c = \mathbf{\tilde{U}}_c \cdot a_c + \mathbf{\hat{U}}_c \cdot b_c\, , \\
        & a_c = \frac{e^{\mathbf{A}_c\mathbf{z}}}{e^{\mathbf{A}_c\mathbf{z}} + e^{\mathbf{B}_c\mathbf{z}}}, \quad b_c = \frac{e^{\mathbf{B}_c\mathbf{z}}}{e^{\mathbf{A}_c\mathbf{z}} + e^{\mathbf{B}_c\mathbf{z}}},  \\
        & \mathbf{U} = \mathbf{\tilde{U}} +  \mathbf{\hat{U}}, \quad s_c = 
        \frac{1}{{N}}\sum_{i=1}^{{N}}\mathbf{U}_c(i), \quad
        \mathbf{z} = \eta(\mathbf{W}\mathbf{s}),
    \end{aligned}
\right .
\end{equation}
where $\mathbf{\hat{U}}, \mathbf{\tilde{U}} \in \mathbb{R}^{N\times C}$ 
are point features encoded by two kernels respectively, 
$\mathbf{\tilde{V}} \in \mathbb{R}^{N\times C}$ is the final fused features,
$\mathbf{s}$ is obtained by using element-wise average pooling over all $N$ points for each feature $c\in C$,
$\eta$ is a FC layer, $\mathbf{W}\in \mathbb{R}^{d\times C}$, $\mathbf{A},\mathbf{B}\in \mathbb{R}^{C\times d}$, and $d$ is a reduced feature size.

Furthermore, we add an residual path besides the main path (shown in Fig.~\ref{fig:point_modules} (c)) and then construct the Residual Point Selective Kernel module (R-PSK) that is used as a building block for RENet.
As shown in Fig.~\ref{fig:spsakn}, RENet follows a hierarchical encoder-decoder architecture by using Edge-preserved Pooling (EP) and Edge-preserved Unpooling (EU) modules~\cite{pan2019pointatrousgraph}. 
We use an Edge-aware Feature Expansion (EFE) module~\cite{pan2020ecg} to expand point features, which generates high-resolution complete point clouds with predicted fine local details.
Consequently, 
multi-scale
structural relations 
can be exploited for fine details generation.

\subsection{Loss Functions}
Our \OM{} 
is trained end-to-end,
and the training loss consists of three parts: $\mathcal{L}_{rec}$ (reconstruction path), 
$\mathcal{L}_{com}$ (completion path) 
and $\mathcal{L}_{fine}$ (relational enhancement).
$\mathcal{L}_{rec}$ and $\mathcal{L}_{com}$ have two loss items, 
a $\mathbf{KL}$ divergence loss and a reconstruction loss, while
$\mathcal{L}_{fine}$ only has a reconstruction loss.
The KL divergence is defined as:
\begin{equation}
    \mathcal{L}_{\mathbf{KL}}(q, p) = -\mathbf{KL} \big[q(\mathbf{z})\,\big\|\,p(\mathbf{z})\big].
\end{equation}
Considering the training efficiency, we choose the symmetric Chamfer Distance (CD) as the reconstruction loss:
\begin{equation}
    \mathcal{L}_{\mathbf{CD}}(\mathbf{P}, \mathbf{Q}) = \frac{1}{|\mathbf{P}|}\sum_{x\in\mathbf{P}}\underset{y\in\mathbf{Q}}{\min}\|x-y\|^2 + \frac{1}{|\mathbf{Q}|}\sum_{y\in\mathbf{Q}}\underset{x\in\mathbf{P}}{\min}\|x-y\|^2,
    \label{eq:cd_loss}
\end{equation}
where $x$ and $y$ denote points that belong to two point clouds
$\mathbf{P}$ 
and $\mathbf{Q}$, respectively.
Consequently, the joint loss function can be formulated as:
\begin{equation}
\begin{aligned}
    \mathcal{L} = & \lambda_{rec}\mathcal{L}_{rec} + \lambda_{com}\mathcal{L}_{com} + \lambda_{fine}\mathcal{L}_{fine} \\
    = &\lambda_{rec}\big[ \mathcal{L}_\mathbf{KL}(q_{\phi}(\mathbf{z_g}|\mathbf{Y}),\, \mathcal{N}(\mathbf{0}, \mathbf{I})) + \mathcal{L}_\mathbf{CD}(\mathbf{Y}_r^{\prime}, \mathbf{Y}) \big] \\
    + &\lambda_{com}\big[ \mathcal{L}_\mathbf{KL}(p_{\psi}(\mathbf{z_g}|\mathbf{X}), \, q_{\phi}(\mathbf{z_g}|\mathbf{Y})) + \mathcal{L}_{\mathbf{CD}}(\mathbf{Y}_c^{\prime}, \mathbf{Y}) \big] \\
    + &\lambda_{fine}\mathcal{L}_\mathbf{CD}(\mathbf{Y}_f^{\prime}, \mathbf{Y}),
\end{aligned}
\end{equation}
where $\lambda_f$, $\lambda_r$ and $\lambda_c$ are the weighting parameters.

{
\setlength{\tabcolsep}{6.3pt}
\begin{table*}
    \caption{Shape completion results (CD loss multiplied by $10^4$) on our multi-view partial point cloud dataset (16,384 points). \OM{} outperforms all existing methods by convincing margins. Note that besides the conventional 8 categories in existing datasets, MVP allows evaluation on 8 additional categories.}
    \vspace{-7mm}
    \begin{center}
    \small{
        \begin{tabular}{l|cccccccc|cccccccc|c}
            \Xhline{1pt}
            {Method} & {\rotatebox{75}{\scriptsize airplane }} & {\rotatebox{75}{\scriptsize cabinet }} & {\rotatebox{75}{\scriptsize car }} & {\rotatebox{75}{\scriptsize chair }} & {\rotatebox{75}{\scriptsize lamp }} & {\rotatebox{75}{\scriptsize sofa }} & {\rotatebox{75}{\scriptsize table }} & {\rotatebox{75}{\scriptsize watercraft }} & {\rotatebox{75}{\scriptsize bed }} & {\rotatebox{75}{\scriptsize bench }} & {\rotatebox{75}{\scriptsize bookshelf }} & {\rotatebox{75}{\scriptsize bus }} & {\rotatebox{75}{\scriptsize guitar }} & {\rotatebox{75}{\scriptsize motorbike }} & {\rotatebox{75}{\scriptsize pistol }} & {\rotatebox{75}{\scriptsize skateboard }} & {Avg.} \\
             \hline\hline
            \scriptsize PCN~\cite{yuan2018pcn} & \scriptsize 2.95 & \scriptsize 4.13 & \scriptsize 3.04 & \scriptsize 7.07 & \scriptsize 14.93 & \scriptsize 5.56 & \scriptsize 7.06 & \scriptsize 6.08 & \scriptsize 12.72 & \scriptsize 5.73 & \scriptsize 6.91 & \scriptsize 2.46 & \scriptsize 1.02 & \scriptsize 3.53 & \scriptsize 3.28 & \scriptsize 2.99 & \scriptsize 6.02 \\
             
            \scriptsize TopNet~\cite{tchapmi2019topnet} & \scriptsize 2.72 & \scriptsize 4.25 & \scriptsize 3.40 & \scriptsize 7.95 & \scriptsize 17.01 & \scriptsize 6.04 & \scriptsize 7.42 & \scriptsize 6.04 & \scriptsize 11.60 & \scriptsize 5.62 & \scriptsize 8.22 & \scriptsize 2.37 & \scriptsize 1.33 & \scriptsize 3.90 & \scriptsize 3.97 & \scriptsize 2.09 & \scriptsize 6.36 \\
             
            \scriptsize MSN~\cite{liu2020morphing} & \scriptsize 2.07 & \scriptsize 3.82 & \scriptsize 2.76 & \scriptsize 6.21 & \scriptsize 12.72 & \scriptsize 4.74 & \scriptsize 5.32 & \scriptsize 4.80 & \scriptsize 9.93 & \scriptsize 3.89 & \scriptsize 5.85 & \scriptsize 2.12 & \scriptsize 0.69 & \scriptsize 2.48 & \scriptsize 2.91 & \scriptsize 1.58 & \scriptsize 4.90 \\
             
            \scriptsize Wang et. al.~\cite{wang2020cascaded} & \scriptsize 1.59 & \scriptsize 3.64 & \scriptsize 2.60 & \scriptsize 5.24 & \scriptsize 9.02 & \scriptsize 4.42 & \scriptsize 5.45 & \scriptsize 4.26 & \scriptsize 9.56 & \scriptsize 3.67 & \scriptsize 5.34 & \scriptsize 2.23 & \scriptsize 0.79 & \scriptsize 2.23 & \scriptsize 2.86 & \scriptsize 2.13 & \scriptsize 4.30 \\
             
            \scriptsize ECG~\cite{pan2020ecg} & \scriptsize 1.41 & \scriptsize 3.44 & \scriptsize 2.36 & \scriptsize 4.58 & \scriptsize 6.95 & \scriptsize 3.81 & \scriptsize 4.27 & \scriptsize 3.38 & \scriptsize 7.46 & \scriptsize 3.10 & \scriptsize 4.82 & \scriptsize 1.99 & \scriptsize 0.59 & \scriptsize 2.05 & \scriptsize 2.31 & \scriptsize 1.66 & \scriptsize 3.58 \\
             
            \scriptsize GRNet~\cite{xie2020grnet} & \scriptsize 1.61 & \scriptsize 4.66 & \scriptsize 3.10 & \scriptsize 4.72 & \scriptsize 5.66 & \scriptsize 4.61 & \scriptsize 4.85 & \scriptsize 3.53 & \scriptsize 7.82 & \scriptsize 2.96 & \scriptsize 4.58 & \scriptsize 2.97 & \scriptsize 1.28 & \scriptsize 2.24 & \scriptsize 2.11 & \scriptsize 1.61 & \scriptsize 3.87 \\
            
            \scriptsize NSFA~\cite{zhang2020detail} & \scriptsize 1.51 & \scriptsize 4.24 & \scriptsize 2.75 & \scriptsize 4.68 & \scriptsize 6.04 & \scriptsize 4.29 & \scriptsize 4.84 & \scriptsize 3.02 & \scriptsize 7.93 & \scriptsize 3.87 & \scriptsize 5.99 & \scriptsize 2.21 & \scriptsize 0.78 & \scriptsize 1.73 & \scriptsize 2.04 & \scriptsize 2.14 & \scriptsize 3.77 \\
            \hline
            
            \scriptsize \OM{} (Ours) & \scriptsize \textbf{1.15} & \scriptsize \textbf{3.20} & \scriptsize \textbf{2.14} & \scriptsize \textbf{3.58} & \scriptsize \textbf{5.57} & \scriptsize \textbf{3.58} & \scriptsize \textbf{4.17} & \scriptsize \textbf{2.47} & \scriptsize \textbf{6.90} & \scriptsize \textbf{2.76} & \scriptsize \textbf{3.45} & \scriptsize \textbf{1.78} & \scriptsize \textbf{0.59} & \scriptsize \textbf{1.52} & \scriptsize \textbf{1.83} & \scriptsize \textbf{1.57} & \scriptsize \textbf{3.06} \\
            
            \Xhline{1pt}
            
        \end{tabular}
    }
    \end{center}
    \label{tab:mvpc_CD_16384}
    \vspace{-5mm}
\end{table*}
}
\section{Multi-View Partial Point Cloud Dataset}~\label{sec:mvp}
Towards an effort to build a more unified and comprehensive dataset for incomplete point clouds,
we contribute the MVP dataset, a high-quality multi-view partial point cloud dataset, to the community.  
We compare the MVP dataset to previous partial point cloud benchmarks, PCN~\cite{yuan2018pcn} and Completion3D~\cite{tchapmi2019topnet} in Table~\ref{tab:mvp_comp}.
The MVP dataset has many advantages over the other datasets.

\noindent\textbf{Diversity \& Uniform Views.}
First, 
the MVP dataset consists of diverse partial point clouds.
Instead of rendering partial shapes by using randomly selected camera poses~\cite{yuan2018pcn,tchapmi2019topnet}, we select 26 camera poses that are uniformly distributed on a unit sphere for each CAD model (Fig.~\ref{fig:mv_all} (a)).
Notably,
the relative poses between our 26 camera poses are fixed, but the first camera pose is randomly selected, which is equivalent to performing a random rotation to all 26 camera poses.
The major advantages of 
using uniformly distributed camera views
are threefold:
\textbf{1)} The MVP dataset has fewer similar rendered partial 3D shapes than the other datasets.
\textbf{2)} Its partial point clouds rendered by uniformly distributed camera views can cover most parts of a complete 3D shape.
\textbf{3)} We can generate sufficient incomplete-complete 3D shape pairs with a relatively small number of 3D CAD models.
According to Tatarchenko et. al.~\cite{tatarchenko2019single}, many 3D reconstruction methods rely primarily on shape recognition; they essentially perform shape retrieval from the massive training data.
Hence, using fewer complete shapes during training can better evaluate the capability of generating complete 3D shapes conditioned on the partial observation, rather than naively retrieving a known similar complete shape.
An example of 26 rendered partial point clouds are shown in Fig.~\ref{fig:mv_all} (b).

\noindent\textbf{Large-Scale \& High-Resolution.}
Second,
the MVP dataset consists of over 100,000 high-quality incomplete and complete point clouds.
Previous methods render incomplete point clouds by using small virtual camera resolutions (\eg 160 $\times$ 120), which is much smaller than real depth cameras (\eg both Kinect V2 and Intel RealSense are 1920 $\times$ 1080). 
Consequently, the rendered partial point clouds are unrealistic.
In contrast, we use the resolution 1600 $\times$ 1200 to render partial 3D shapes of high quality (Fig.~\ref{fig:mv_all} (c)).
For ground truth, 
we employ Poisson Disk Sampling (PDS)~\cite{bridson2007fast,kazhdan2013screened} to sample non-overlapped and uniformly spaced points for complete shapes (Fig.~\ref{fig:mv_all} (d)). PDS yields smoother complete point clouds than uniform sampling, making them a better representation of the underlying object CAD models. Hence, we can better evaluate network capabilities of recovering high-quality geometric details.
Previous datasets provide complete shapes with only one resolution.
Unlike those datasets, we create complete point clouds with different resolutions, including 2048(1x), 4096(2x), 8192(4x) and 16384(8x) for precisely evaluating the completion quality at different resolutions.

{
\setlength{\tabcolsep}{5.95pt}
\begin{table*}
    \caption{Shape completion results (F-Score@1\%) on our 
    multi-view partial (MVP) point cloud dataset (16,384 points). 
    }
    \vspace{-7.5mm}
    \begin{center}
        \small{
            \begin{tabular}{l|cccccccc|cccccccc|c}
            \Xhline{1pt}
            {Method} & {\rotatebox{75}{\scriptsize airplane}} & {\rotatebox{75}{\scriptsize cabinet }} & {\rotatebox{75}{\scriptsize car }} & {\rotatebox{75}{\scriptsize chair }} & {\rotatebox{75}{\scriptsize lamp }} & {\rotatebox{75}{\scriptsize sofa }} & {\rotatebox{75}{\scriptsize table }} & {\rotatebox{75}{\scriptsize watercraft }} & {\rotatebox{75}{\scriptsize bed }} & {\rotatebox{75}{\scriptsize bench }} & {\rotatebox{75}{\scriptsize bookshelf }} & {\rotatebox{75}{\scriptsize bus }} & {\rotatebox{75}{\scriptsize guitar }} & {\rotatebox{75}{\scriptsize motorbike }} & {\rotatebox{75}{\scriptsize pistol }} & {\rotatebox{75}{\scriptsize skateboard }} & {Avg.} \\
             \hline\hline
            \scriptsize PCN~\cite{yuan2018pcn} & \fontsize{6}{6}\selectfont 0.816 & \fontsize{6}{6}\selectfont 0.614 & \fontsize{6}{6}\selectfont 0.686 & \fontsize{6}{6}\selectfont 0.517 & \fontsize{6}{6}\selectfont 0.455 & \fontsize{6}{6}\selectfont 0.552 & \fontsize{6}{6}\selectfont 0.646 & \fontsize{6}{6}\selectfont 0.628 & \fontsize{6}{6}\selectfont 0.452 & \fontsize{6}{6}\selectfont 0.694 & \fontsize{6}{6}\selectfont 0.546 & \fontsize{6}{6}\selectfont 0.779 & \fontsize{6}{6}\selectfont 0.906 & \fontsize{6}{6}\selectfont 0.665 & \fontsize{6}{6}\selectfont 0.774 & \fontsize{6}{6}\selectfont 0.861 & \fontsize{6}{6}\selectfont 0.638 \\
             
            \scriptsize TopNet~\cite{tchapmi2019topnet} & \fontsize{6}{6}\selectfont 0.789 & \fontsize{6}{6}\selectfont 0.621 & \fontsize{6}{6}\selectfont 0.612 & \fontsize{6}{6}\selectfont 0.443 & \fontsize{6}{6}\selectfont 0.387 & \fontsize{6}{6}\selectfont 0.506 & \fontsize{6}{6}\selectfont 0.639 & \fontsize{6}{6}\selectfont 0.609 & \fontsize{6}{6}\selectfont 0.405 & \fontsize{6}{6}\selectfont 0.680 & \fontsize{6}{6}\selectfont 0.524 & \fontsize{6}{6}\selectfont 0.766 & \fontsize{6}{6}\selectfont 0.868 & \fontsize{6}{6}\selectfont 0.619 & \fontsize{6}{6}\selectfont 0.726 & \fontsize{6}{6}\selectfont 0.837 & \fontsize{6}{6}\selectfont 0.601 \\
             
            \scriptsize MSN~\cite{liu2020morphing} & \fontsize{6}{6}\selectfont 0.879 & \fontsize{6}{6}\selectfont 0.692 & \fontsize{6}{6}\selectfont 0.693 & \fontsize{6}{6}\selectfont 0.599 & \fontsize{6}{6}\selectfont 0.604 & \fontsize{6}{6}\selectfont 0.627 & \fontsize{6}{6}\selectfont 0.730 & \fontsize{6}{6}\selectfont 0.696 & \fontsize{6}{6}\selectfont 0.569 & \fontsize{6}{6}\selectfont 0.797 & \fontsize{6}{6}\selectfont 0.637 & \fontsize{6}{6}\selectfont 0.806 & \fontsize{6}{6}\selectfont 0.935 & \fontsize{6}{6}\selectfont 0.728 & \fontsize{6}{6}\selectfont 0.809 & \fontsize{6}{6}\selectfont 0.885 & \fontsize{6}{6}\selectfont 0.710 \\
             
            \scriptsize Wang et. al.~\cite{wang2020cascaded} & \fontsize{6}{6}\selectfont 0.898 & \fontsize{6}{6}\selectfont 0.688 & \fontsize{6}{6}\selectfont 0.725 & \fontsize{6}{6}\selectfont 0.670 & \fontsize{6}{6}\selectfont 0.681 & \fontsize{6}{6}\selectfont 0.641 & \fontsize{6}{6}\selectfont 0.748 & \fontsize{6}{6}\selectfont 0.742 & \fontsize{6}{6}\selectfont 0.600 & \fontsize{6}{6}\selectfont 0.797 & \fontsize{6}{6}\selectfont 0.659 & \fontsize{6}{6}\selectfont 0.802 & \fontsize{6}{6}\selectfont 0.931 & \fontsize{6}{6}\selectfont 0.772 & \fontsize{6}{6}\selectfont 0.843 & \fontsize{6}{6}\selectfont 0.902 & \fontsize{6}{6}\selectfont 0.740 \\
             
            \scriptsize ECG~\cite{pan2020ecg} & \fontsize{6}{6}\selectfont 0.906 & \fontsize{6}{6}\selectfont 0.680 & \fontsize{6}{6}\selectfont 0.716 & \fontsize{6}{6}\selectfont 0.683 & \fontsize{6}{6}\selectfont 0.734 & \fontsize{6}{6}\selectfont 0.651 & \fontsize{6}{6}\selectfont 0.766 & \fontsize{6}{6}\selectfont 0.753 & \fontsize{6}{6}\selectfont 0.640 & \fontsize{6}{6}\selectfont 0.822 & \fontsize{6}{6}\selectfont 0.706 & \fontsize{6}{6}\selectfont 0.804 & \fontsize{6}{6}\selectfont 0.945 & \fontsize{6}{6}\selectfont 0.780 & \fontsize{6}{6}\selectfont 0.835 & \fontsize{6}{6}\selectfont 0.897 & \fontsize{6}{6}\selectfont 0.753 \\
             
            \scriptsize GRNet~\cite{xie2020grnet} & \fontsize{6}{6}\selectfont 0.853 & \fontsize{6}{6}\selectfont 0.578 & \fontsize{6}{6}\selectfont 0.646 & \fontsize{6}{6}\selectfont 0.635 & \fontsize{6}{6}\selectfont 0.710 & \fontsize{6}{6}\selectfont 0.580 & \fontsize{6}{6}\selectfont 0.690 & \fontsize{6}{6}\selectfont 0.723 & \fontsize{6}{6}\selectfont 0.586 & \fontsize{6}{6}\selectfont 0.765 & \fontsize{6}{6}\selectfont 0.635 & \fontsize{6}{6}\selectfont 0.682 & \fontsize{6}{6}\selectfont 0.865 & \fontsize{6}{6}\selectfont 0.736 & \fontsize{6}{6}\selectfont 0.787 & \fontsize{6}{6}\selectfont 0.850 & \fontsize{6}{6}\selectfont 0.692 \\
            
            \scriptsize NSFA~\cite{zhang2020detail} & \fontsize{6}{6}\selectfont 0.903 & \fontsize{6}{6}\selectfont 0.694 & \fontsize{6}{6}\selectfont 0.721 & \fontsize{6}{6}\selectfont 0.737 & \fontsize{6}{6}\selectfont 0.783 & \fontsize{6}{6}\selectfont \textbf{0.705} & \fontsize{6}{6}\selectfont \textbf{0.817} & \fontsize{6}{6}\selectfont 0.799 & \fontsize{6}{6}\selectfont \textbf{0.687} & \fontsize{6}{6}\selectfont 0.845 & \fontsize{6}{6}\selectfont 0.747 & \fontsize{6}{6}\selectfont 0.815 & \fontsize{6}{6}\selectfont 0.932 & \fontsize{6}{6}\selectfont 0.815 & \fontsize{6}{6}\selectfont 0.858 & \fontsize{6}{6}\selectfont 0.894 & \fontsize{6}{6}\selectfont 0.783 \\
            \hline
            
            \scriptsize \OM{} (Ours) & \fontsize{6}{6}\selectfont \textbf{0.928} & \fontsize{6}{6}\selectfont \textbf{0.721} & \fontsize{6}{6}\selectfont \textbf{0.756} & \fontsize{6}{6}\selectfont \textbf{0.743} & \fontsize{6}{6}\selectfont \textbf{0.789} & \fontsize{6}{6}\selectfont 0.696 & \fontsize{6}{6}\selectfont 0.813 & \fontsize{6}{6}\selectfont \textbf{0.800} & \fontsize{6}{6}\selectfont 0.674 & \fontsize{6}{6}\selectfont \textbf{0.863} & \fontsize{6}{6}\selectfont \textbf{0.755} & \fontsize{6}{6}\selectfont \textbf{0.832} & \fontsize{6}{6}\selectfont \textbf{0.960} & \fontsize{6}{6}\selectfont \textbf{0.834} & \fontsize{6}{6}\selectfont \textbf{0.887} & \fontsize{6}{6}\selectfont \textbf{0.930} & \fontsize{6}{6}\selectfont \textbf{0.796} \\
            \Xhline{1pt}
            
        \end{tabular}
        }
    \end{center}
    \label{tab:mvpc_F1_16384}
    \vspace{-5mm}
\end{table*}
}

\noindent\textbf{Rich Categories.}
Third,
the MVP dataset consists of 16 shape categories of partial and complete shapes for training and testing.
Besides the 8 categories (airplane, cabinet, car, chair, lamp, sofa, table and watercraft) included in previous datasets~\cite{yuan2018pcn,tchapmi2019topnet}, we add another 8 categories (bed, bench, bookshelf, bus, guitar, motorbike, pistol and skateboard).
By using more categories of shapes, it becomes more challenging to train and evaluate networks on the MVP dataset.

To sum up, our MVP dataset consists of a large number of 
high-quality
synthetic
partial scans for 3D CAD models, which imitates real-scanned incomplete point clouds caused by self-occlusion.
Besides 3D shape completion, our MVP dataset can be used in many other partial point cloud tasks, such as classification, registration and keypoints extraction.
Compared to previous partial point cloud datasets, MVP dataset has many favorable properties.
More detailed comparisons between our dataset and previous datasets are reported in our supplementary materials.



\section{Experiments}
\noindent\textbf{Evaluation Metrics.}
In line with previous methods~\cite{tchapmi2019topnet,xie2020grnet,zhang2020detail}, we evaluate the reconstruction accuracy by computing the Chamfer Distance (Eq.~\eqref{eq:cd_loss}) between the predicted complete shapes $\mathbf{Y}^{\prime}$ and the ground truth shapes $\mathbf{Y}$.
Based on the insight that CD can be misleading due to its sensitivity to outliers~\cite{tatarchenko2019single}, we also use F-score~\cite{knapitsch2017tanks} to evaluate the distance between object surfaces, which
is defined as the harmonic mean between precision and recall.

\noindent\textbf{Implementation Details.}
Our networks are implemented using PyTorch.
We train our models using the Adam optimizer~\cite{kingma2014adam} with initial learning rate 1e$^{-4}$ (decayed by 0.7 every 40 epochs) and batch size 32 by NVIDIA TITAN Xp GPU.
Note that \OM{} does not use any symmetry tricks, such as reflection symmetry or mirror operations.

\subsection{Shape Completion on Our MVP Dataset}

\noindent\textbf{Quantitative Evaluation.}
As introduced in Sec.~\ref{sec:mvp}, our MVP dataset consists of 16 categories of high-quality partial/complete point clouds that are generated by CAD models selected from the ShapeNet~\cite{wu20153d} dataset.
We split our models into a training set (62,400 shape pairs) and a test set (41,600 shape pairs).
Note 
that none of the complete shapes in our test set are included in our training set.
To achieve a fair comparison, we train all methods using the same training strategy on our MVP dataset.
The evaluated CD loss and F-score for all evaluated methods (16,384 points) are reported in Table~\ref{tab:mvpc_CD_16384} and Table~\ref{tab:mvpc_F1_16384}, respectively.
\OM{} outperforms all existing competitive methods in terms of CD and F-score@1\%.
Moreover, \OM{} can generate complete point clouds with various resolutions ($N=$ 2048, 4096, 8192 and 16384). 
We compare our methods with existing approaches that support multi-resolution completion in Table~\ref{tab:res}, and \OM{} outperforms all the other methods.
\setlength{\tabcolsep}{4.85pt}
\begin{table}
    \caption{Shape completion results (CD loss multiplied by $10^4$) with various resolutions. 
    }
    \vspace{-7mm}
    \begin{center}
    \small{
        \begin{tabular}{l|cc|cc|cc|cc}
            \Xhline{1pt}
            \multirow{2}{*}{\# Points} & \multicolumn{2}{c|}{2,048} & \multicolumn{2}{c|}{4,096} & \multicolumn{2}{c|}{8,192} & \multicolumn{2}{c}{16,384} \\
            \cline{2-9}
             & \scriptsize CD & \scriptsize F1 & \scriptsize CD & \scriptsize F1 & \scriptsize CD & \scriptsize F1 & \scriptsize CD & \scriptsize F1  \\
            \hline\hline
            \scriptsize PCN~\cite{yuan2018pcn} & \fontsize{6}{6}\selectfont 9.77 & \fontsize{6}{6}\selectfont 0.320 & \fontsize{6}{6}\selectfont 7.96 & \fontsize{6}{6}\selectfont 0.458 & \fontsize{6}{6}\selectfont 6.99 & \fontsize{6}{6}\selectfont 0.563 & \fontsize{6}{6}\selectfont 6.02 & \fontsize{6}{6}\selectfont 0.638\\
            \scriptsize TopNet~\cite{tchapmi2019topnet} & \fontsize{6}{6}\selectfont 10.11 & \fontsize{6}{6}\selectfont 0.308 & \fontsize{6}{6}\selectfont 8.20 & \fontsize{6}{6}\selectfont 0.440 & \fontsize{6}{6}\selectfont 7.00 & \fontsize{6}{6}\selectfont 0.533 & \fontsize{6}{6}\selectfont 6.36 & \fontsize{6}{6}\selectfont 0.601 \\
            \scriptsize MSN~\cite{liu2020morphing} & \fontsize{6}{6}\selectfont 7.90 & \fontsize{6}{6}\selectfont 0.432 & \fontsize{6}{6}\selectfont 6.17 & \fontsize{6}{6}\selectfont 0.585 & \fontsize{6}{6}\selectfont 5.42 & \fontsize{6}{6}\selectfont 0.659 & \fontsize{6}{6}\selectfont 4.90 & \fontsize{6}{6}\selectfont 0.710 \\
            \scriptsize Wang et. al.~\cite{wang2020cascaded} & \fontsize{6}{6}\selectfont 7.25 & \fontsize{6}{6}\selectfont 0.434 & \fontsize{6}{6}\selectfont 5.83 & \fontsize{6}{6}\selectfont 0.569 & \fontsize{6}{6}\selectfont 4.90 & \fontsize{6}{6}\selectfont 0.680 & \fontsize{6}{6}\selectfont 4.30 & \fontsize{6}{6}\selectfont 0.740 \\
            \scriptsize ECG~\cite{pan2020ecg} & \fontsize{6}{6}\selectfont 6.64 & \fontsize{6}{6}\selectfont 0.476 & \fontsize{6}{6}\selectfont 5.41 & \fontsize{6}{6}\selectfont 0.585 & \fontsize{6}{6}\selectfont 4.18 & \fontsize{6}{6}\selectfont 0.690 & \fontsize{6}{6}\selectfont 3.58 & \fontsize{6}{6}\selectfont 0.753 \\
            \hline
            \scriptsize \OM{} (Ours) & \fontsize{6}{6}\selectfont \textbf{5.96} & \fontsize{6}{6}\selectfont \textbf{0.499} & \fontsize{6}{6}\selectfont \textbf{4.70} & \fontsize{6}{6}\selectfont \textbf{0.636} & \fontsize{6}{6}\selectfont \textbf{3.64} & \fontsize{6}{6}\selectfont \textbf{0.727} & \fontsize{6}{6}\selectfont \textbf{3.12} & \fontsize{6}{6}\selectfont \textbf{0.791}\\
            \Xhline{1pt}
        \end{tabular}
    }
    \end{center}
    \vspace{-4mm}
    \label{tab:res}
\end{table}

\begin{figure*}
    \centering
    \includegraphics[width=1\linewidth]{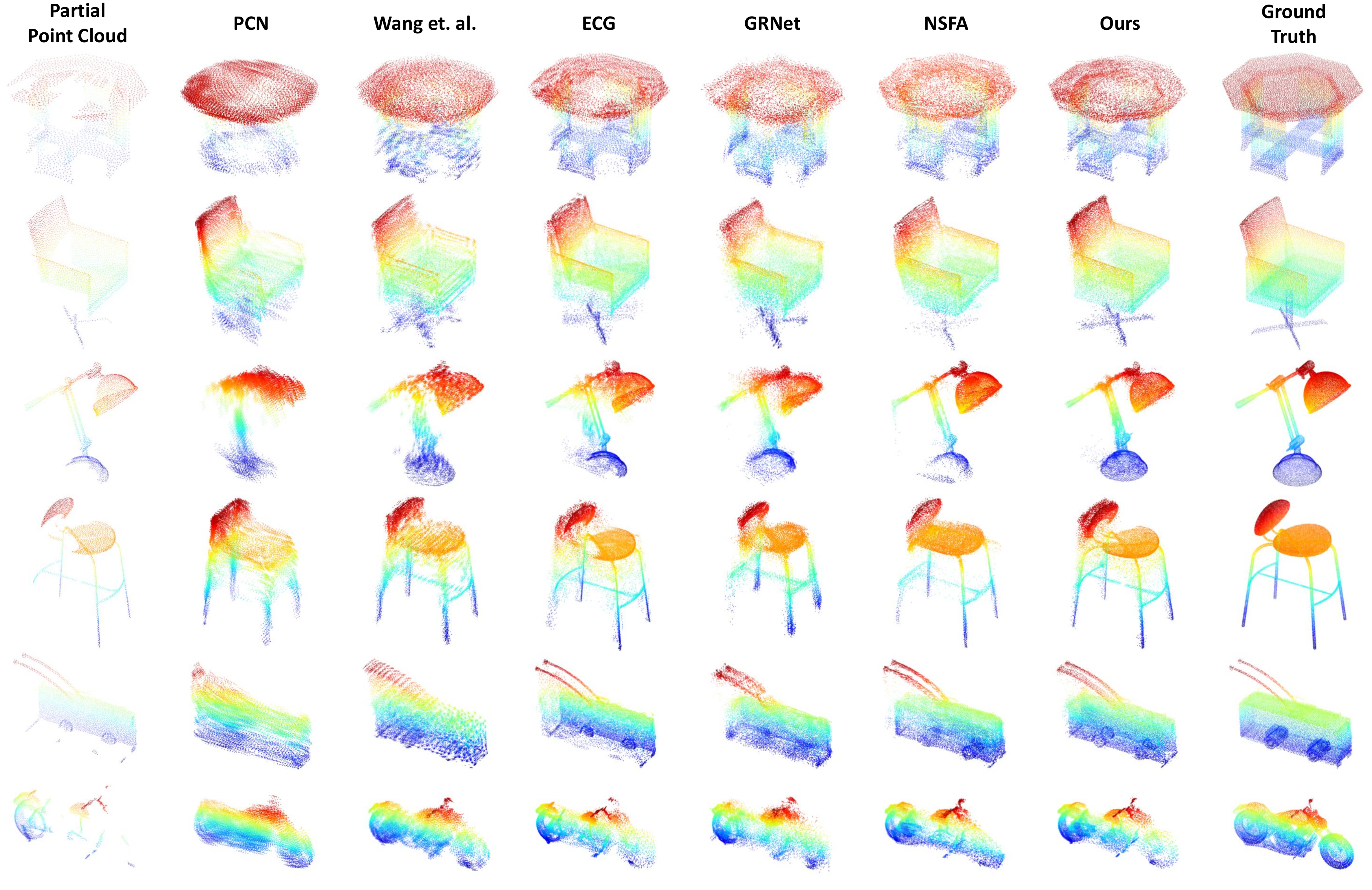}
    \vspace{-7.75mm}
    \caption{Qualitative completion results (16,384 points) on the MVP dataset by different methods. \OM{} can generate better complete point clouds than the other methods by learning geometrical symmetries.}
    \vspace{-3mm}
    \label{fig:qua_all}
\end{figure*}

{
\setlength{\tabcolsep}{7.3pt}
\begin{table}[]
    \caption{Ablation studies (2,048 points) for the proposed network modules, including Point Self-Attention Kernel, Dual-path Architecture and Point Selective Kernel Module.}
    \vspace{-7mm}
    \begin{center}
        \small{
        \begin{tabular}{ccc|cc}
            \Xhline{1pt}
            \makecell{Point \\ Self-Attention} & \makecell{Dual-path \\ Architecture} & \makecell{Kernel \\ Selection} & \scriptsize CD & \scriptsize F1 \\
            \hline\hline
             & & & \footnotesize 6.64 & \footnotesize 0.476  \\
            \checkmark & & & \footnotesize 6.35 & \footnotesize 0.484  \\
            \checkmark & \checkmark & & \footnotesize 6.15 & \footnotesize 0.492 \\
            \checkmark & \checkmark & \checkmark & \footnotesize 5.96 & \footnotesize 0.499 \\
            \Xhline{1pt}
        \end{tabular}
        }
    \end{center}
    \vspace{-4mm}
    \label{tab:ablation}
\end{table}
}
\noindent\textbf{Qualitative Evaluation.}
The qualitative comparison results are shown in Fig.~\ref{fig:qua_all}.
The proposed \OM{} can generate better complete shapes with fine details than the other methods.
In particular, 
we can clearly observe the learned relational structures in our complete shapes.
For example, the missing legs of the chairs (the second row and the fourth row in Fig.~\ref{fig:qua_all}) are recovered based on the observed legs with the learned shape symmetry.
In the third row of Fig.~\ref{fig:qua_all}, we reconstruct the incomplete lamp base with a smooth round bowl shape, which makes it a more plausible completion than the others.
The partially observed motorbike in the last row does not contain its front wheel, and \OM{} reconstructs a complete wheel by learning the observed back wheel.
Consequently, \OM{} can effectively reconstruct complete shapes by learning structural relations, including geometrical symmetries, regular arrangements and surface smoothness, from the incomplete point cloud.

\noindent\textbf{Ablation Study.}
The ablation studies 
for
all our proposed modules, Point Self-Attention Kernel (PSA), Dual-path Architecture and Kernel Selection (two-branch PSK), 
are presented in Table~\ref{tab:ablation}.
We use ECG~\cite{pan2020ecg} as our baseline model and evaluate the completion results with 2048 points.
By adding the proposed modules, better completion results can be achieved, which validates their effectiveness.

\subsection{Shape Completion on Completion3D}
The Completion3D benchmark is an online platform for evaluating 3D shape completion approaches.
Following their instructions, we train \OM{} using their prepared training data and upload our best completion results (2,048 points). 
As reported in the online leaderboard{\href{https://completion3d.stanford.edu/results}{$^1$}}, also shown in Table~\ref{tab:c3D_CD_2048}, \OM{} significantly outperforms SOTA methods and is ranked first on the Completion3D benchmark.

\setlength{\tabcolsep}{5pt}
\begin{table}
    \caption{Shape completion results (CD loss multiplied by $10^4$) on the Completion3D benchmark (2,048 points). Our \OM{} outperforms all SOTAs by significant margins.}
    \vspace{-7mm}
    \begin{center}
    \small{
        \begin{tabular}{l|cccccccc|c}
            \Xhline{1pt}
            \scriptsize{Method} & {\rotatebox{75}{\fontsize{5}{5}\selectfont airplane }} & {\rotatebox{75}{\fontsize{5}{5}\selectfont cabinet }} & {\rotatebox{75}{\fontsize{5}{5}\selectfont car }} & {\rotatebox{75}{\fontsize{5}{5}\selectfont chair }} & {\rotatebox{75}{\fontsize{5}{5}\selectfont lamp }} & {\rotatebox{75}{\fontsize{5}{5}\selectfont sofa }} & {\rotatebox{75}{\fontsize{5}{5}\selectfont table }} & {\rotatebox{75}{\fontsize{5}{5}\selectfont watercraft }} & \scriptsize{Avg.} \\
             \hline\hline
            \fontsize{5}{5}\selectfont AtlasNet~\cite{groueix2018atlasnet} & \fontsize{5}{5}\selectfont 10.36 & \fontsize{5}{5}\selectfont 23.40 & \fontsize{5}{5}\selectfont 13.40 & \fontsize{5}{5}\selectfont 24.16 & \fontsize{5}{5}\selectfont 20.24 & \fontsize{5}{5}\selectfont 20.82 & \fontsize{5}{5}\selectfont 17.52 & \fontsize{5}{5}\selectfont 11.62 & \fontsize{5}{5}\selectfont 17.77 \\ 
            
            \fontsize{5}{5}\selectfont PCN~\cite{yuan2018pcn} & \fontsize{5}{5}\selectfont 9.79 & \fontsize{5}{5}\selectfont 22.70 & \fontsize{5}{5}\selectfont 12.43 & \fontsize{5}{5}\selectfont 25.14 & \fontsize{5}{5}\selectfont 22.72 & \fontsize{5}{5}\selectfont 20.26 & \fontsize{5}{5}\selectfont 20.27 & \fontsize{5}{5}\selectfont 11.73 & \fontsize{5}{5}\selectfont 18.22 \\ 
             
            \fontsize{5}{5}\selectfont TopNet~\cite{tchapmi2019topnet} & \fontsize{5}{5}\selectfont 7.32 & \fontsize{5}{5}\selectfont 18.77 & \fontsize{5}{5}\selectfont 12.88 & \fontsize{5}{5}\selectfont 19.82 & \fontsize{5}{5}\selectfont 14.60 & \fontsize{5}{5}\selectfont 16.29 & \fontsize{5}{5}\selectfont 14.89 & \fontsize{5}{5}\selectfont 8.82 & \fontsize{5}{5}\selectfont 14.25 \\ 
             
            \fontsize{5}{5}\selectfont GRNet~\cite{xie2020grnet} & \fontsize{5}{5}\selectfont 6.13 & \fontsize{5}{5}\selectfont 16.90 & \fontsize{5}{5}\selectfont 8.27 & \fontsize{5}{5}\selectfont 12.23 & \fontsize{5}{5}\selectfont 10.22 & \fontsize{5}{5}\selectfont 14.93 & \fontsize{5}{5}\selectfont 10.08 & \fontsize{5}{5}\selectfont 5.86 & \fontsize{5}{5}\selectfont 10.64 \\
            
            \hline
            
            \fontsize{5}{5}\selectfont \OM{} (Ours) & \fontsize{5}{5}\selectfont \textbf{3.94} & \fontsize{5}{5}\selectfont \textbf{10.93} & \fontsize{5}{5}\selectfont \textbf{6.44} & \fontsize{5}{5}\selectfont \textbf{9.32} & \fontsize{5}{5}\selectfont \textbf{8.32} & \fontsize{5}{5}\selectfont \textbf{11.35} & \fontsize{5}{5}\selectfont \textbf{8.60} & \fontsize{5}{5}\selectfont \textbf{5.78} & \fontsize{5}{5}\selectfont \textbf{8.12} \\
            \Xhline{1pt}
            
        \end{tabular}
    }    
    \end{center}
    \vspace{-4mm}
    \label{tab:c3D_CD_2048}
\end{table}
\subsection{Shape Completion on Real-world Partial Scans}
\begin{figure}[t!]
    \centering
    \includegraphics[width=1\linewidth]{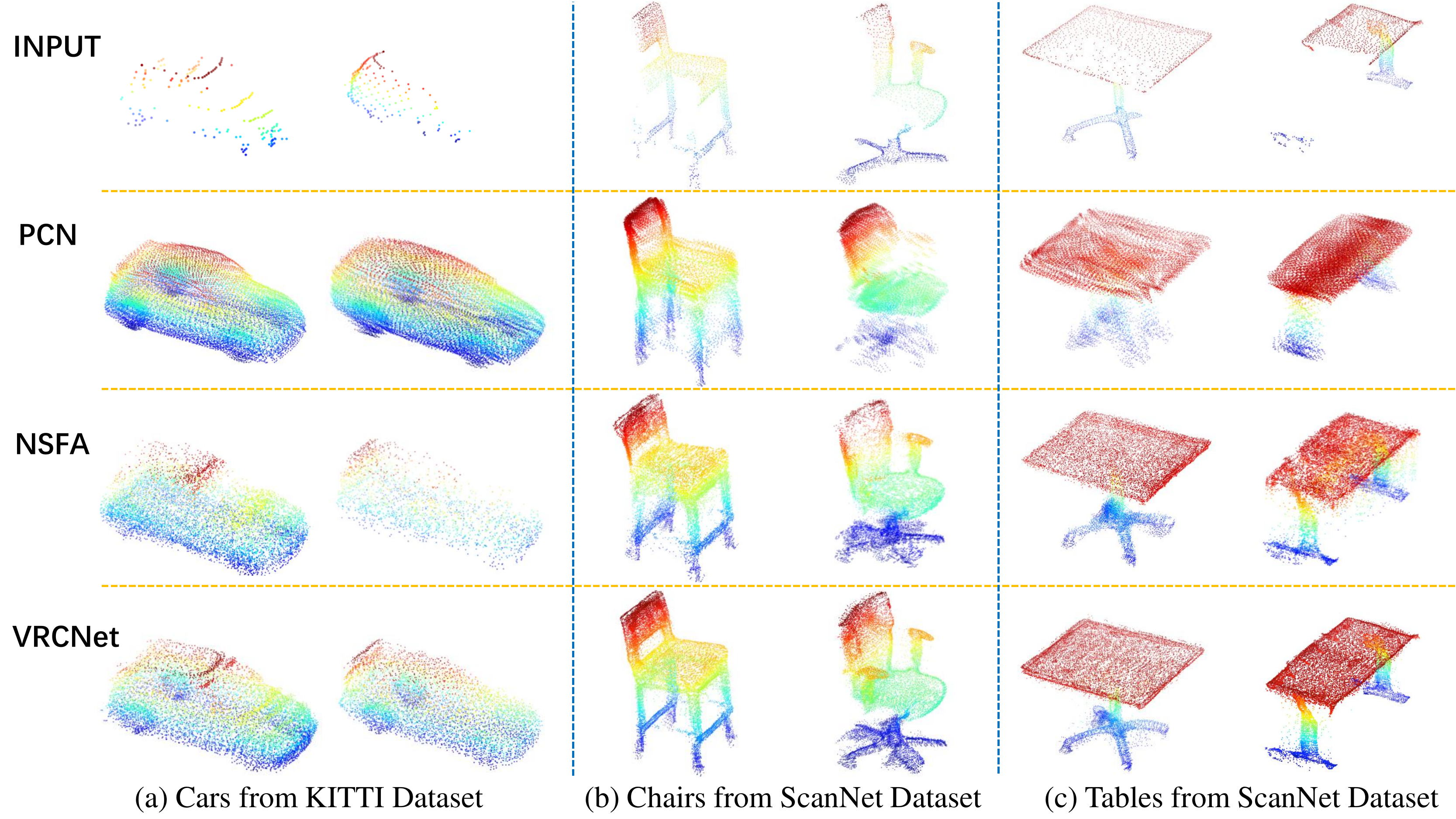}
    \vspace{-6mm}
    \caption{
    \OM{} generates impressive complete shapes for real-scanned point clouds by learning and predicting shape symmetries. 
    (a) shows completion results for cars from Kitti dataset~\cite{Geiger2012CVPR}. 
    (b) and (c) show completion results for chairs and tables from ScanNet dataset~\cite{dai2017scannet}, respectively.}
    \label{fig:real_scan}
    \vspace{-3mm}
\end{figure}

We further evaluate \OM{} (trained on MVP with all categories) on real scans, including 
cars from the KITTI~\cite{Geiger2012CVPR} dataset, chairs and tables from the ScanNet dataset~\cite{dai2017scannet}.
It is noteworthy
that the KITTI dataset captured point clouds by using a LiDAR whereas the ScanNet dataset uses a depth camera.
For sparse LiDAR data, we fine-tune all trained models on ShapeNet-car dataset, but no fine-tuning is needed for chairs and tables.
The qualitative completion results are shown in Fig.~\ref{fig:real_scan}.
For those sparse point clouds of cars, \OM{} can predict complete and smooth surfaces that also preserves the observed shape details.  
In comparison, PCN~\cite{yuan2018pcn} suffers a loss of fine shape details and NSFA~\cite{zhang2020detail} cannot generate high-quality complete shapes due to large missing ratios.
For those incomplete chairs and tables, \OM{} generates appealing complete point clouds by exploiting the shape symmetries in the partial scans.


\noindent\textbf{User Study.}
We conduct a user study on completion results for real-scanned point clouds by PCN, NSFA and \OM{}, where our \OM{} is the most preferred method overall. More details are reported in our supplementary materials.

\section{Conclusion}
In this paper, we propose \OM{}, a variational relational point completion network, which effectively exploits 3D structural relations to predict complete shapes.
Novel self-attention modules, such as 
PSA and PSK,
are proposed for adaptively learning point cloud features, which can be conveniently used in other point cloud tasks.
In addition, we contribute a large-scale MVP dataset, which consists of over 100,000 high-quality 3D point clouds.
We highly encourage researchers to use our proposed novel modules and the MVP dataset for future studies on partial point clouds.

\section*{Acknowledgement}
This research was conducted in collaboration with SenseTime. This work is supported by NTU NAP and A*STAR through the Industry Alignment Fund - Industry Collaboration Projects Grant.

{\small
\bibliographystyle{ieee_fullname}
\bibliography{main}
}

\clearpage
\begin{center}
    \Large{\textbf{Supplementary Material}}
\end{center}

\appendix

\section{Overview}
In this supplementary material, we provide 
in-depth method analysis (Sec.~\ref{sec:ana}), 
inference details (Sec.~\ref{sec:inf}), 
detailed dataset comparisons (Sec.~\ref{sec:dc}), 
comprehensive ablation studies (Sec.~\ref{sec:as}),
resource usages (Sec.~\ref{sec:rs}),
and 
the user study on real scans (Sec.~\ref{sec:us}). 
Qualitative results with different settings are shown in the corresponding sections.

\section{Analysis} \label{sec:ana}
\subsection{Variational Modeling}
\begin{figure}
    \centering
    \includegraphics[width=1\linewidth]{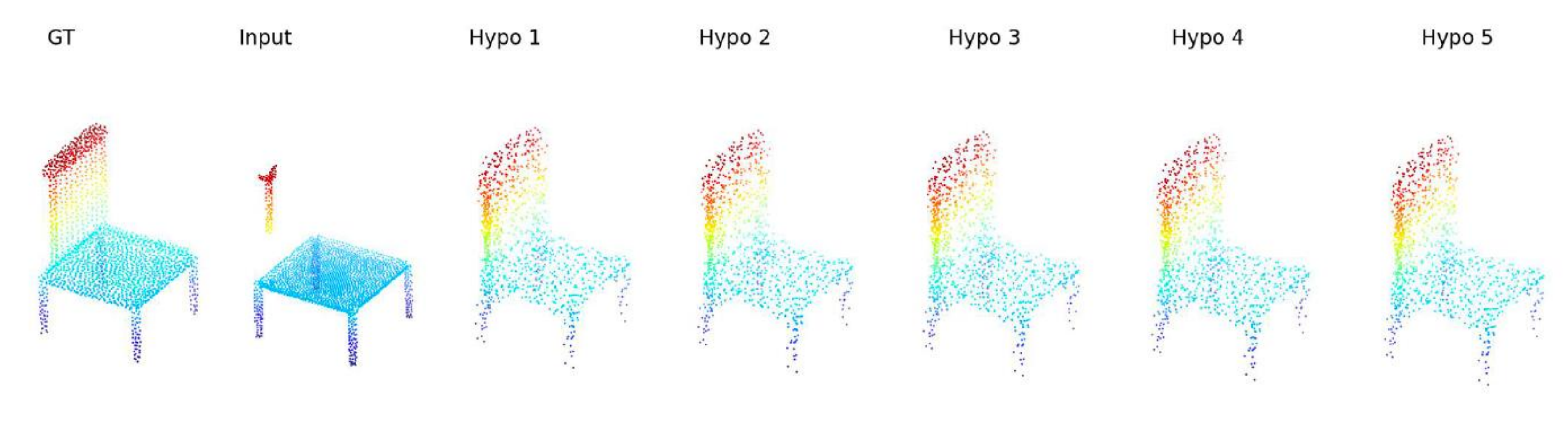}
    \vspace{-10mm}
    \caption{\textbf{Coarse Completion Results by 
    Different Samples.} We can observe that the generated different hypotheses for coarse shape skeletons are similar with each other.}
    \label{fig:vm}
\end{figure}

Inspired by~\cite{zheng2019pluralistic}, our VMNet consists of two parallel paths: 1) a reconstruction path, and 2) a completion path.
Both two paths follow similar variational auto-encoder structures, which generates complete point clouds using embedded global features and predicted distributions.
During training, the encoded distributions (posterior) for incomplete point clouds (completion path) are guided by the encoded distributions (prior) for complete point clouds (reconstruction path).
In this way, we mitigate the domain gap between the posterior and the prior distributions by regularizing the posterior distribution to approach the prior distribution. 
Consequently, the learned smooth complete shape priors are incorporated into our shape completion process. 
During inference, we only use the completion path.
We randomly generate a sample from the learned posterior distribution $p_{\psi}(\mathbf{z_g}|\mathbf{X})$ for shape completion.
Theoretically, diverse plausible coarse completions can be generated by using different samples from $p_{\psi}(\mathbf{z_g}|\mathbf{X})$.
However, we observe similar predicted coarse completions for different samples (see Fig.~\ref{fig:vm}).
In other words, different samples do not influence our completion results.
According to~\cite{zheng2019pluralistic}, employing a generative adversarial learning scheme can highly increase the shape completion diversity,
which we leave as a future research direction.

\subsection{Local Point Relation Learning}
\begin{figure}
    \centering
    \includegraphics[width=1\linewidth]{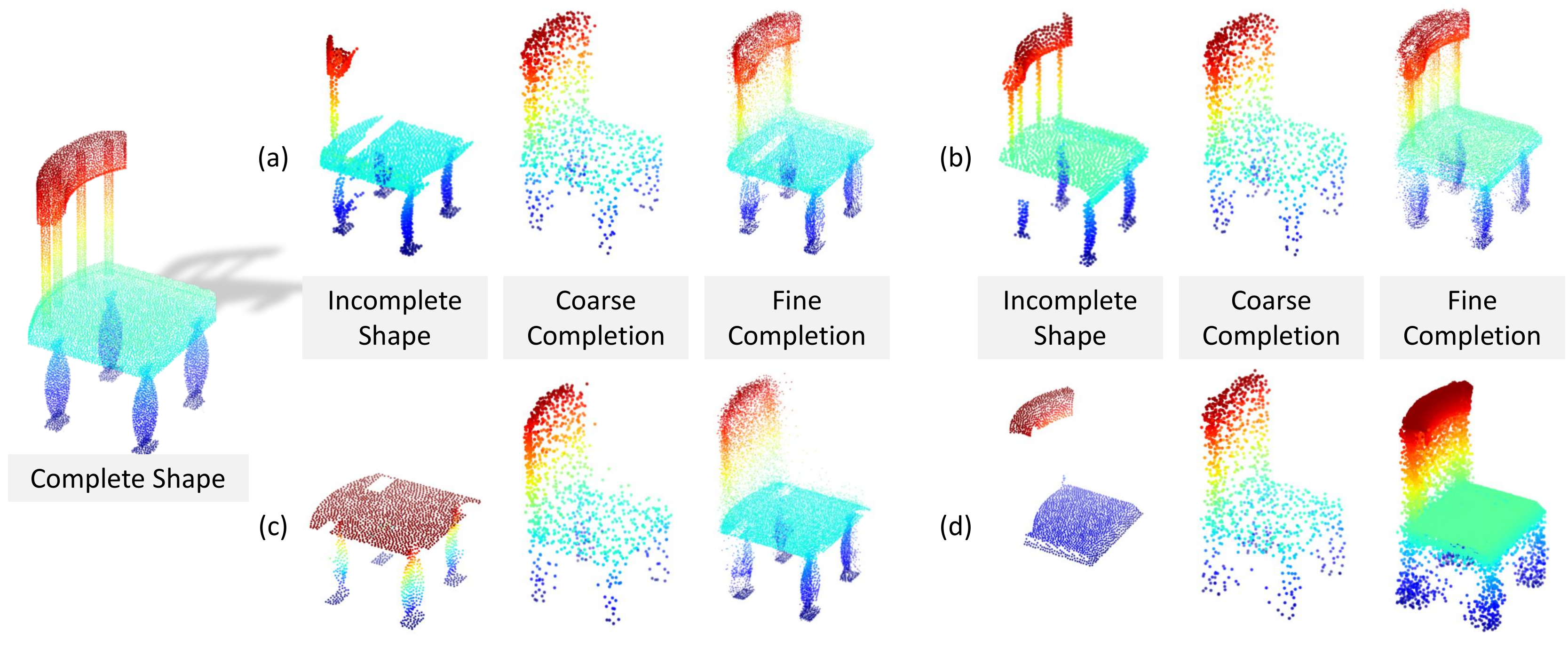}
    \vspace{-7mm}
    \caption{\textbf{Observing different parts of the chair, the \OM{} generates different complete chairs based on the partial observations and predicted shape skeletons.}  In (a) and (b), the \OM{} predicts complete chair shapes by learning the shape geometrical symmetry from the partial observations.  Both (c) and (d) show the incomplete shapes with large missing ratios, and the \OM{} predicts the fine complete shapes based on the coarse complete shapes.}
    \label{fig:mv_chair}
\end{figure}

\begin{figure*}
    \centering
    \includegraphics[width=1\linewidth]{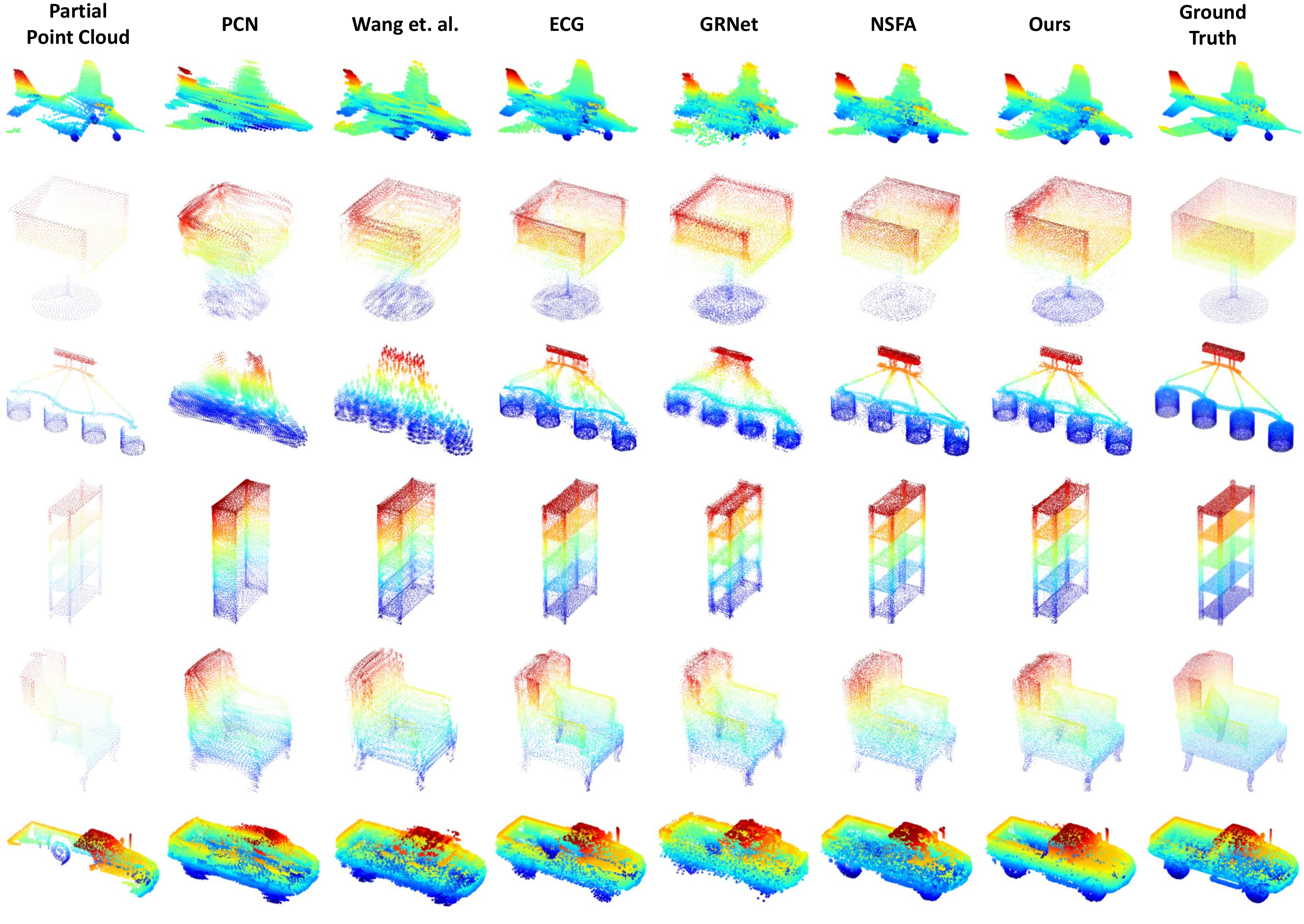}
    \caption{\textbf{Qualitative Results on the MVP Dataset by Different Methods.} Note that we use different point sizes for different shapes to achieve better visualizations.}
    \label{fig:qua_all_supp}
\end{figure*}

\begin{figure*}
    \centering
    \includegraphics[width=1\linewidth]{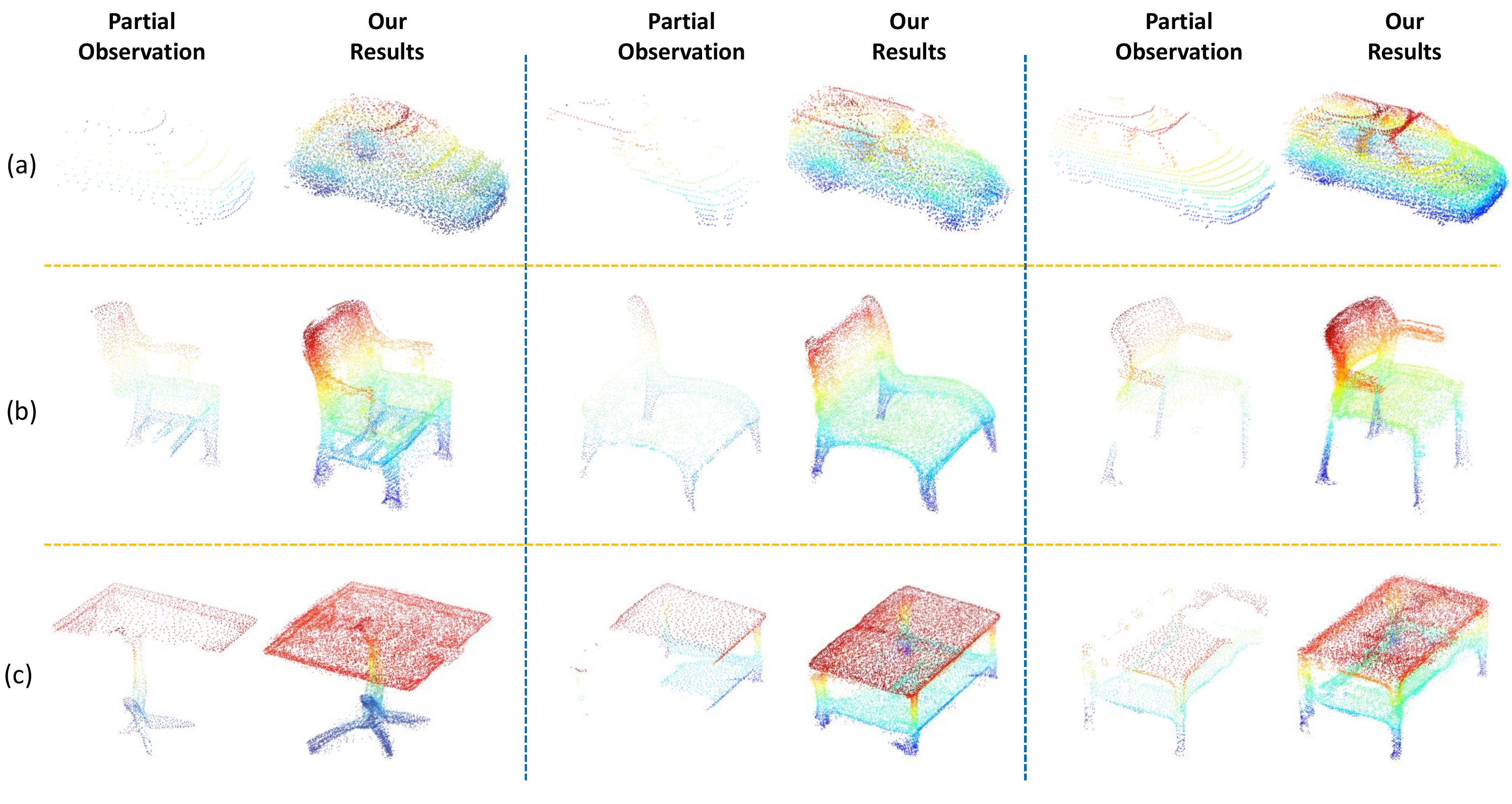}
    \caption{\textbf{Our completion results for real-scanned incomplete point clouds.} (a) shows our completion results for incomplete cars from the KITTI dataset~\cite{Geiger2012CVPR}. (b) and (c) show our results for scanned incomplete chairs and tables, respectively.}
    \label{fig:real_supp}
\end{figure*}

Relation operations~\cite{hu2019local,zhao2020exploring} (also known as self-attention operations)  adaptively learn a meaningful compositional structure that is used to predict adaptive weights by exploiting relations among local elements.
Comparing to conventional convolution operations that use fixed weights, relation operations adapt aggregation weights based on the composability of local elements.
Motivated by the success of using relation operations in natural language process and image applications, we expand and use relation operations to learn point relations in neighboring points for point cloud completion.
Previous methods~\cite{zhang2020detail,pan2020ecg,xie2020grnet} preserve observed local details from the incomplete point clouds in their completion results by learning local point features.
However, they cannot generate fine-grained complete shapes for those missing parts.
Consequently, their completion results often have high-quality observed shape parts and low-quality missing shape parts.
In contrast, with the help of self-attention operations, our RENet can adaptively recover fine-grained complete shapes by implicitly predicting shape structural relations, such as geometrical symmetries, regular arrangements and surface smoothness.
More qualitative results by different methods are shown in Fig.~\ref{fig:qua_all_supp}, which show that the RENet can effectively learn structural relations for shape completion.
For example, the lamp completion results (the 3rd row of Fig.~\ref{fig:qua_all_supp}) show that the \OM{} can recover those cylinder bulbs by the learned geometrical symmetry.
In particular, given different observed incomplete shapes, the \OM{} can generate different complete 3D shapes with the help of both structural relations from the partial observations and the generated coarse overall shape skeletons (shown in Fig.~\ref{fig:mv_chair}).
Moreover, the \OM{} can generate pluralistic complete shapes for real-scanned incomplete point clouds (see Fig.~\ref{fig:real_supp}), which validates its strong robustness and generaliability.


\section{Inference Details} \label{sec:inf}
Our \OM{} consists of two consecutive sub-networks, PMNet and RENet.
PMNet generates overall shape skeletons (i.e. coarse completions) using probabilistic modeling, 
and RENet enhances structural relations at multiple scales to generate our fine completions.
For inference, PMNet only uses its completion path to predict coarse completions based on the incomplete point clouds.
A coarse complete point cloud that consists of 1024 points can be regarded as 3D adaptive points to facilitate learning local point relations. 
The coarse completion is combined with the incomplete point cloud (2048 points) as the input (3072 points) to the RENet.
After exploiting multi-scale point features, RENet uses the Edge-aware Feature Expansion (EFE) module~\cite{pan2020ecg} to upsample and expand the point feature so as to generate complete point clouds with different resolutions.
For example, we generate complete shapes with 16384 points by 1) upsampling to 18432 points (3072 \texttimes{} 5) and thus 2) downsampling to 16384 points using farthest point sampling.
\begin{table}[t]
\caption{A user study of completion quality on real scans. The values are average scores given by volunteers (3 points for best result, 1 point for the worst result). \OM{} is the most preferred method overall}
\vspace{-5mm}
\begin{center}
\begin{tabular}{l|c|c|c}
\Xhline{1pt}
Category & PCN~\cite{yuan2018pcn} & NSFA~\cite{zhang2020detail} & \OM \\
\hline\hline
Car (KITTI)      & \textbf{2.87} & 1.07 & 2.07  \\
Chair (ScanNet)   & 1.60  & 1.73 &\textbf{2.67}  \\
Table (ScanNet)    & 1.27  & 2.20 &\textbf{2.60}   \\
\hline
Overall & 1.91 & 1.67 & \textbf{2.45} \\
\Xhline{1pt}
\end{tabular}
\end{center}
\label{tab:user_study}
\end{table}
\section{User Study on Real Scans} \label{sec:us}
We conduct a user study on the performances of various methods in Tab \ref{tab:user_study}. 
Specifically, we gather a group of 15 volunteers to rank the quality  of complete point cloud predicted by PCN, NSFA, and our \OM, on the real scans of three object categories: car, chair and table. 
For each object category, the volunteers are given three anonymous groups of results, produced by three methods. 
The volunteers are instructed to give the best, middle, and worst results 3, 2, and 1 point(s) respectively. 
We then compute the average scores of all volunteers for each method and class category. 
The evaluation is conducted in a double-blind manner (the methods are anonymous to both the instructor and the volunteers) and the order of the groups are shuffled for each category. 
Our \OM{} is the most favored method overall amongst the three. 
PCN obtains higher score for car completion because it generates smooth mean shapes for all cars, even though few observed shape details of those cars are preserved in their completion results.
For the other two categories, chair and table, the \OM{} receives the highest scores due to its effectiveness on reconstructing complete shapes using predicted shape symmetries.
{
\setlength{\tabcolsep}{7.3pt}
\begin{table}[]
    \vspace{-3mm}
    \caption{Ablation studies (2,048 points) for the proposed network modules, including Point Self-Attention Kernel, Dual-Path Architecture and Point Selective Kernel Module.}
    \vspace{-7mm}
    \begin{center}
        \small{
        \begin{tabular}{ccc|cc}
            \Xhline{1pt}
            \makecell{Point \\ Self-Attention} & \makecell{Dual-Path \\ Architecture} & \makecell{Kernel \\ Selection} & \scriptsize CD & \scriptsize F1 \\
            \hline\hline
             & & & \footnotesize 6.64 & \footnotesize 0.476  \\
             & \checkmark & & \footnotesize 6.43 & \footnotesize 0.488  \\
            \checkmark & & & \footnotesize 6.35 & \footnotesize 0.484  \\
             & \checkmark & \checkmark & \footnotesize 6.35 & \footnotesize 0.490  \\
            \checkmark & \checkmark & & \footnotesize 6.15 & \footnotesize 0.492 \\
            \checkmark & \checkmark & \checkmark & \footnotesize 5.96 & \footnotesize 0.499 \\
            \Xhline{1pt}
        \end{tabular}
        }
    \end{center}
    \vspace{-3mm}
    \label{tab:ablation_all}
\end{table}
}
\setlength{\tabcolsep}{4pt}
\begin{table*}[]
    \caption{\textbf{Comparing MVP with existing datasets.} 
    MVP has many appealing properties, such as
    1) diversity of uniform views; 2) large-scale and high-quality; 3) rich categories. 
    Note that both PCN and C3D only randomly render \textbf{One} incomplete point cloud for each CAD model to construct their testing sets.
    (C3D: Completion3D; Cat.: Categories; Distri.: Distribution; Reso.: Resolution; PC: Point Cloud; FPS: Farthest Point Sampling; PDS: Poisson Disk Sampling. Point cloud resolution is shown as multiples of 2048 points.)}
    \vspace{-7mm}
    \begin{center}
    \small{
        \begin{tabular}{l|c|cc|cc|ccc|cc|cc}
            \Xhline{1pt}
             \multirow{2}{*}{} &
             \multirow{2}{*}{\#Cat.} & 
             \multicolumn{2}{c|}{Training Set} &
             \multicolumn{2}{c}{Testing Set} &
             \multicolumn{3}{|c|}{Virtual Camera} & 
             \multicolumn{2}{c|}{Complete PC} &
             \multicolumn{2}{c}{Incomplete PC} \\
             & & \#CAD & \#Pair & \#CAD & \#Pair & Num. & Distri. & Reso. & 
             Sampling & Reso. & Sampling & Reso. \\
            \hline\hline 
             \small PCN~\cite{yuan2018pcn} & \small 8 & \small 28974 & \small $\sim$200k & \small 1200 & \small 1200 & \small 8 & \small Random & \small {160\texttimes120} & \small Uniform & \small 8\texttimes & \small Random & \small $\sim$3000 \\ 
             \small C3D~\cite{tchapmi2019topnet} & \small 8 & 28974 & \small 28974 & \small 1184 & \small 1184 & \small 1 & \small Random & \small {160\texttimes120} & \small Uniform & \small 1\texttimes & \small Random & \small 1\texttimes \\
            MSN~\cite{liu2020morphing} & \small 8 & \small 28974 & \small $\sim$1.4m & \small 1200 & \small 1200 & \small 50 & \small Random & \small {160\texttimes120} & \small Uniform & \small 4\texttimes & \small Random & \small $\sim$5000 \\
            Wang et. al.~\cite{wang2020cascaded} & \small 8 & \small 28974 & \small 28974 & \small 1200 & \small 1200 & \small 1 & \small Random & \small {160\texttimes120} & \small Uniform & \small 1\texttimes & \small Random & \small 1\texttimes \\
            SANet~\cite{wen2020point} & \small 8 & \small 28974 & \small $\sim$200k & \small 1200 & \small 1200 & \small 8 & \small Random & \small {160\texttimes120} & \small Uniform & \small 1\texttimes & \small Random & \small 1\texttimes \\
            NSFA~\cite{zhang2020detail} & \small 8 & \small 28974 & \small $\sim$200k & \small 1200 & \small 1200 & \small 7 & \small Random & \small {160\texttimes120} & \small Uniform & \small 8\texttimes & \small Random & \small 1\texttimes \\
            \hline
             MVP & \small \textbf{16} & \small 2400 & \small  62400  & \small 1600 & \textbf{41600} & \small \textbf{26} & \small \textbf{Uniform} & \small \textbf{{1600\texttimes1200}} & \small \textbf{PDS} & \small \textbf{1/2/4/8\texttimes} & 
             \small \textbf{FPS} & \small 1\texttimes \\
            \Xhline{1pt}
        \end{tabular}
    }
    \end{center}
    \vspace{-3mm}
    \label{tab:mvp_comp_all}
\end{table*}

\begin{figure*}
    \centering
    \includegraphics[width=1\linewidth]{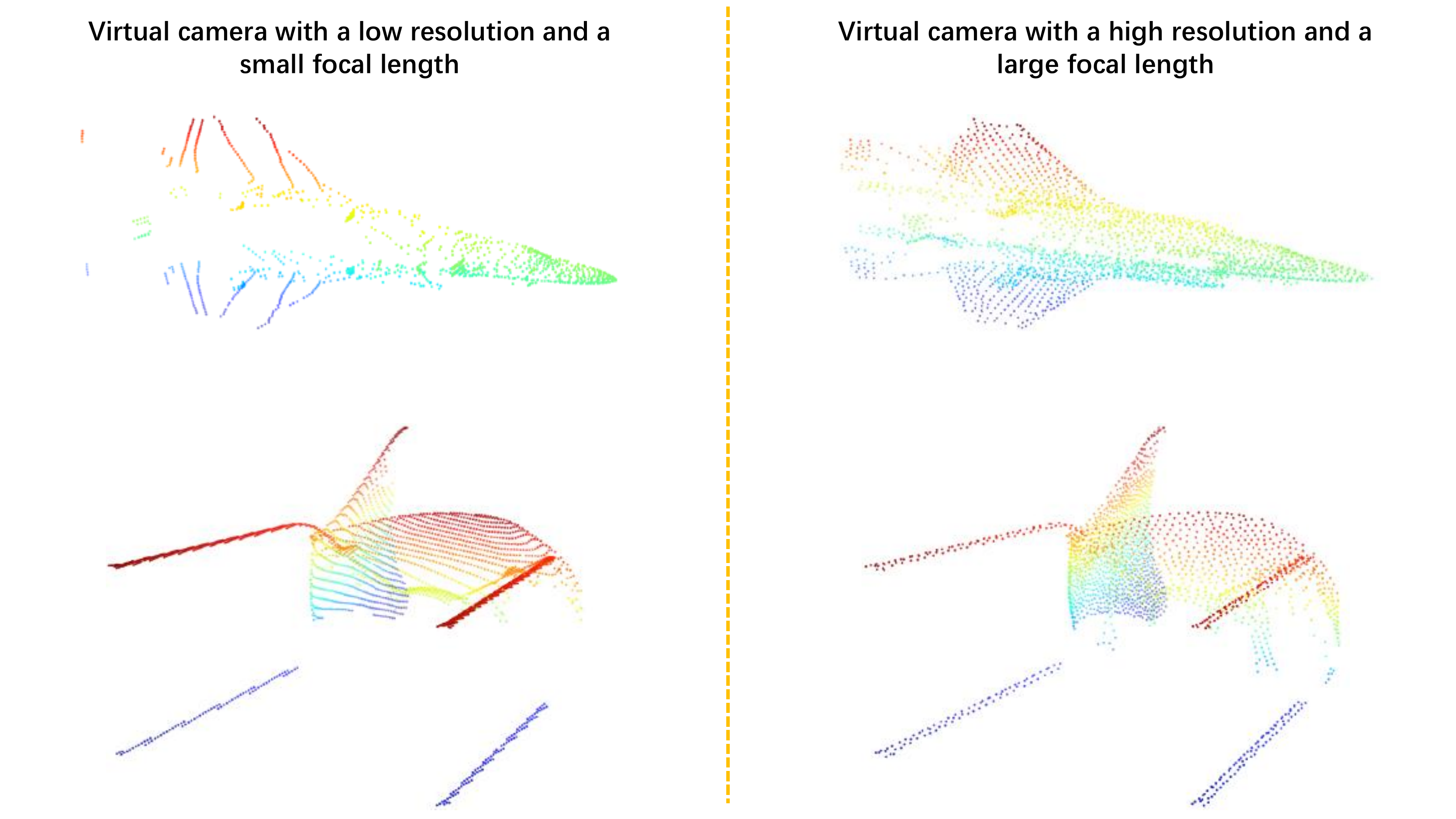}
    \caption{\textbf{Rendered Incomplete Point Clouds with Different Camera Resolutions.} We use a high camera resolution to capture more realistic shapes than using low resolutions.}
    \label{fig:high_low}
\end{figure*}
\begin{figure*}
    \centering
    \includegraphics[width=1\linewidth]{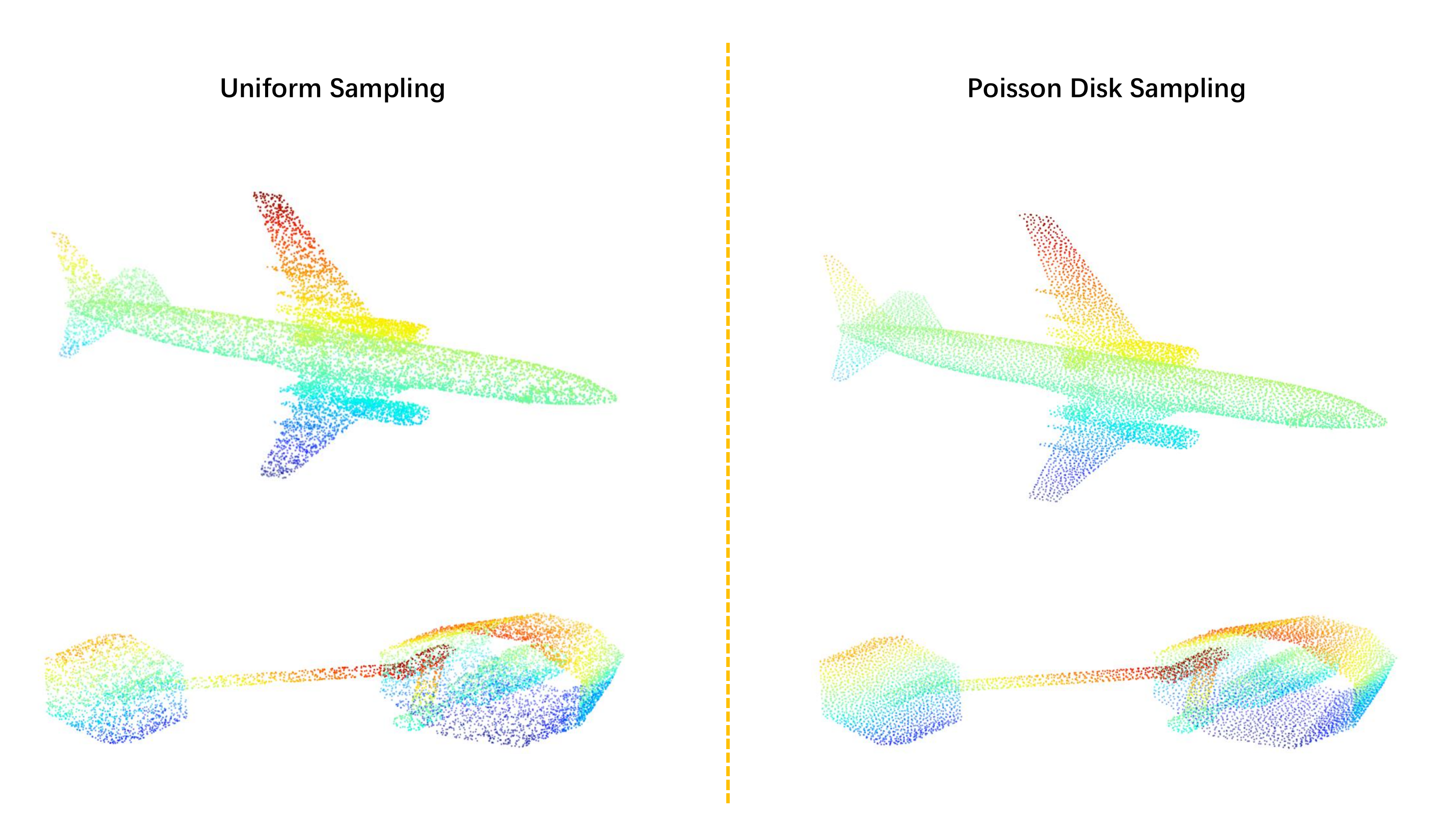}
    \caption{\textbf{Sampling Complete Point Clouds with Different Sampling Methods.} Unlike previous methods that use uniform sampling, we use Poisson Disk Sampling to generate complete point clouds, which can better describe the underlying 3D shape surfaces.}
    \label{fig:pds_us}
\end{figure*}
\section{Ablation Studies} \label{sec:as}
We provide the detailed ablation studies in Table~\ref{tab:ablation_all}, which reports the evaluation results with different combinations of the proposed modules, Point Self-Attention Kernel (PSA), Dual-Path Architecture (DP) and Point Selective Kernel (PSK) Module.
As reported in Table~\ref{tab:ablation_all}, it is obvious that using the proposed modules can improve the completion accuracy.
During training and evaluation, the effectiveness of using PSA and DP are very straightforward.
PSK may lead to fluctuating evaluation results during training, but it can highly improve the point cloud completion performance.

\section{Resource Usage} \label{sec:rs}
We report the resource usages by PCN~\cite{yuan2018pcn}, NSFA~\cite{zhang2020detail} and our \OM{} in the Table~\ref{tab:resource}.
PCN and our \OM{} are implemented using pytorch, and we use the official implementation (by tensorflow) for NSFA.
To achieve a fair comparison for the inference (Inf.) time, we use the same batch size 32 and test all methods by using an NVIDIA V100 GPU on the same workstation.
Note that, NSFA has many non-trainable operations (such as ball query, grouping and sampling), and hence it takes the longest inference time although it has the least parameters.
Our \OM{} achieves significant improvements in completion qualities with an acceptable increment in the computational cost.
\setlength{\tabcolsep}{3.75pt}
\begin{table}
    \caption{Resource Usages.}
    \vspace{-5mm}
    \begin{center}
        \small{
        \begin{tabular}{l|c|c|c}
            \Xhline{1pt}
            Method & \#Params. (M) & Model Size (Mib) & Inf. Time (ms) \\
            \hline\hline
            PCN [29] & 6.86 & 26 & 2.7 \\
            NSFA [30] & 5.38 & 64 & 764.9 \\
            VRCNet & 17.47 & 67 & 183.3 \\
            \Xhline{1pt}
        \end{tabular}
        }
    \end{center}
    \label{tab:resource}
\end{table}
\section{Dataset Comparisons} \label{sec:dc}
As stated in the main paper, previous methods usually use two datasets for incomplete point cloud: ShapeNet~\cite{wu20153d} by PCN~\cite{yuan2018pcn} and Completion3D~\cite{tchapmi2019topnet}.
Because the incomplete point cloud dataset created by PCN is too massive, most following works (including the Completion3D) use a subset of ShapeNet~\cite{wu20153d} derived from PCN~\cite{yuan2018pcn}.
However, 
they do not have a unified and standardized dataset setting, which makes it difficult in directly comparing their performance.
Furthermore, their generated shapes (incomplete and complete point clouds) all have low qualities, which makes their data unrealistic.
In view of this, we create the Multiple-View Partial point cloud (MVP) dataset, which can be a high-quality and unified benchmark for partial point clouds.
The detailed comparisons between different datasets are reported in Table~\ref{tab:mvp_comp_all}.
The proposed MVP dataset has more shape categories (16 v.s. 8), more testing data (e.g. 41600 v.s. 1200) and higher quality point clouds by using better data preparation methods (e.g. PDS, FPS and Uniformly distributed camera poses) than previous datasets.
The qualitative comparisons for rendered incomplete shapes are visualized in Fig.~\ref{fig:high_low}.
By using a high resolution and a large focal length, our rendered partial point clouds are more realistic than using low resolutions and small focal lengths.
Moreover, the qualitative comparisons for sampled complete point clouds by different sampling methods are shown in Fig~\ref{fig:pds_us}.
The MVP dataset uses the Poisson Disk Sampling (PDS) method, which can yield smoother complete point clouds than using uniform sampling.

\href{https://www.youtube.com/watch?v=0SNHlxvCP0g}{\textbf{More qualitative comparisons can be found in our supplementary video.}}


\end{document}